# Enhancing Cryptocurrency Market Forecasting: Advanced Machine Learning Techniques and Industrial Engineering Contributions

Jannatun Nayeem Pinky[*†] and Ramya Akula[1†]

[1]Department of Computer Science

[*]MS in FinTech

[†]University of Central Florida, USA.

**Abstract**

Cryptocurrencies, as decentralized digital assets, have experienced rapid growth and adoption, with over 23,000 cryptocurrencies and a market capitalization nearing $1.1 trillion (about $3,400 per person in the US) as of 2023. This dynamic market presents significant opportunities and risks, highlighting the need for accurate price prediction models to manage volatility. This chapter comprehensively reviews machine learning (ML) techniques applied to cryptocurrency price prediction from 2014 to 2024. We explore various ML algorithms, including linear models, tree-based approaches, and advanced deep learning architectures such as transformers and large language models. Additionally, we examine the role of sentiment analysis in capturing market sentiment from textual data like social media posts and news articles to anticipate price fluctuations. With expertise in optimizing complex systems and processes, industrial engineers are pivotal in enhancing these models. They contribute by applying principles of process optimization, efficiency, and risk mitigation to improve computational performance and data management. This chapter highlights the evolving landscape of cryptocurrency price prediction, the integration of emerging technologies, and the significant role of industrial engineers in refining predictive models. By addressing current limitations and exploring future research directions, this chapter aims to advance the development of more accurate and robust prediction systems, supporting better-informed investment decisions and more stable market behavior.

## 1 Introduction

Cryptocurrencies are digital assets used to exchange value, secure transactions using cryptography, manage the generation of new units, and confirm asset transfers. Unlike government-regulated fiat currencies, cryptocurrencies run on decentralized blockchain networks distributed ledgers that record transactions across several computers (Giudici, Milne, and Vinogradov 2020; Swan 2015).

The notion of digital currencies dates to the 1980s and 1990s, when pioneers such as David Chaum attempted to develop digital currency systems like DigiCash. Early digital currency attempts experienced hurdles such as circulation restrictions, scalability issues, security concerns, and regulatory impediments, limiting their broad acceptance (Iwai 1995). The modern age of cryptocurrencies began with the release of Bitcoin in 2008, launched by an anonymous figure known as Satoshi Nakamoto. Bitcoin was created in reaction to the 2008 global fiscal crisis, intending to create an electronic payment system without the need for central banks or intermediaries (Nakamoto 2008). Nakamoto envisioned a platform that enables fast and cost-effective transactions, eliminating the need for trusted third parties like traditional banks (Bação, et al. 2018). Powered by blockchain technology—a transparent, unchangeable database of transactions—Bitcoin and related cryptocurrencies such as Ethereum provided new features. Ethereum, founded in 2015, pioneered practical smart contracts, enabling developers to design decentralized apps (DApps) with specific rules for ownership and transactions using Turing-complete programming (Buterin 2013). This invention accelerated the rise of decentralized finance (DeFi) and blockchain applications in several industries (Timuçin and Biroğul 2023). Bitcoin's success led to the birth of other cryptocurrencies (altcoins) like Litecoin, Dash, and Ripple, each trying to improve on certain areas of Bitcoin's functionality, such as mining efficiency, transaction speed, and anonymity (Ciaian and Rajcaniova 2018).

As of 2023, the cryptocurrency industry has around 23,000 distinct cryptocurrencies with a total market capitalization of almost $1.1 trillion (Grujić and Lakić n.d.). This quick evolution demonstrates the dynamic and inventive character of the Bitcoin market. The cryptocurrency market's rapid growth and volatility present significant profit potential and risks for traders. Bitcoin's dominance impacts Altcoin behavior, necessitating a reliable price prediction system to manage market dynamics efficiently. This predictive strategy is critical for minimizing risks and increasing trading profits (Wimalagunaratne and Poravi 2018).

Industrial engineers, known for their expertise in optimizing complex systems and processes, have a significant role to play in cryptocurrency. Their efficiency improvement, risk mitigation, and systems analysis skills can contribute to developing more robust and scalable systems for cryptocurrency trading. By integrating principles of operations research and optimization with machine learning techniques, industrial engineers can help design better models for managing cryptocurrency market dynamics, leading to enhanced decision-making, and trading strategies.

The cryptocurrency market's recent meteoric expansion has piqued the interest of investors and experts alike. Nonetheless, inherent volatility remains a big concern. As a result, there is an urgent demand for Bitcoin price prediction and sentiment analysis algorithms. These models can assist investors in making more informed decisions by offering insights into probable price fluctuations and market sentiment. Accurate prediction models may reduce risks, improve trading strategies, and help ensure more stable and predictable market behavior.

Machine learning (ML) algorithms provide powerful tools for tackling these challenges. Industrial engineers, by leveraging their expertise in optimization and process design, can enhance the implementation of these ML systems to forecast future price movements more effectively. By analyzing large volumes of past data and identifying significant patterns, ML systems can produce more accurate predictions. Sentiment analysis methods, which evaluate textual data from social media, news, and other sources, can help us understand market sentiment and its influence on prices. This chapter explores the potential of machine learning (ML) for cryptocurrency price prediction and sentiment analysis, focusing on providing important insights to investors and highlighting the contributions of industrial engineers in enhancing these models.

This chapter investigates a wide range of ML algorithms applied to cryptocurrency price prediction from 2014 to 2024. These include linear models like linear regression and support vector machines (SVMs), tree-based models like random forests and XGBoost, probabilistic models like Naive Bayes, clustering models like Latent Dirichlet Allocation (LDA), and advanced deep learning architectures like the Long Short-Term Memory (LSTM) network, transformers, large language models (LLMs), etc. Industrial engineers can contribute by refining these models and applying their process optimization techniques to improve computational efficiency. Furthermore, sentiment analysis methods, including natural language processing (NLP) approaches, are tested for their ability to capture market sentiment and anticipate price changes. The following sections provide an extensive overview of the theoretical foundations of each model category, emphasizing their strengths and limitations in the context of cryptocurrency prediction. We then delve into the practical applications of these models, discussing the evaluation methods used to assess their effectiveness. The chapter also highlights industrial engineers' role in addressing ML-based prediction challenges and proposes future research topics to improve model performance and incorporate market dynamics.

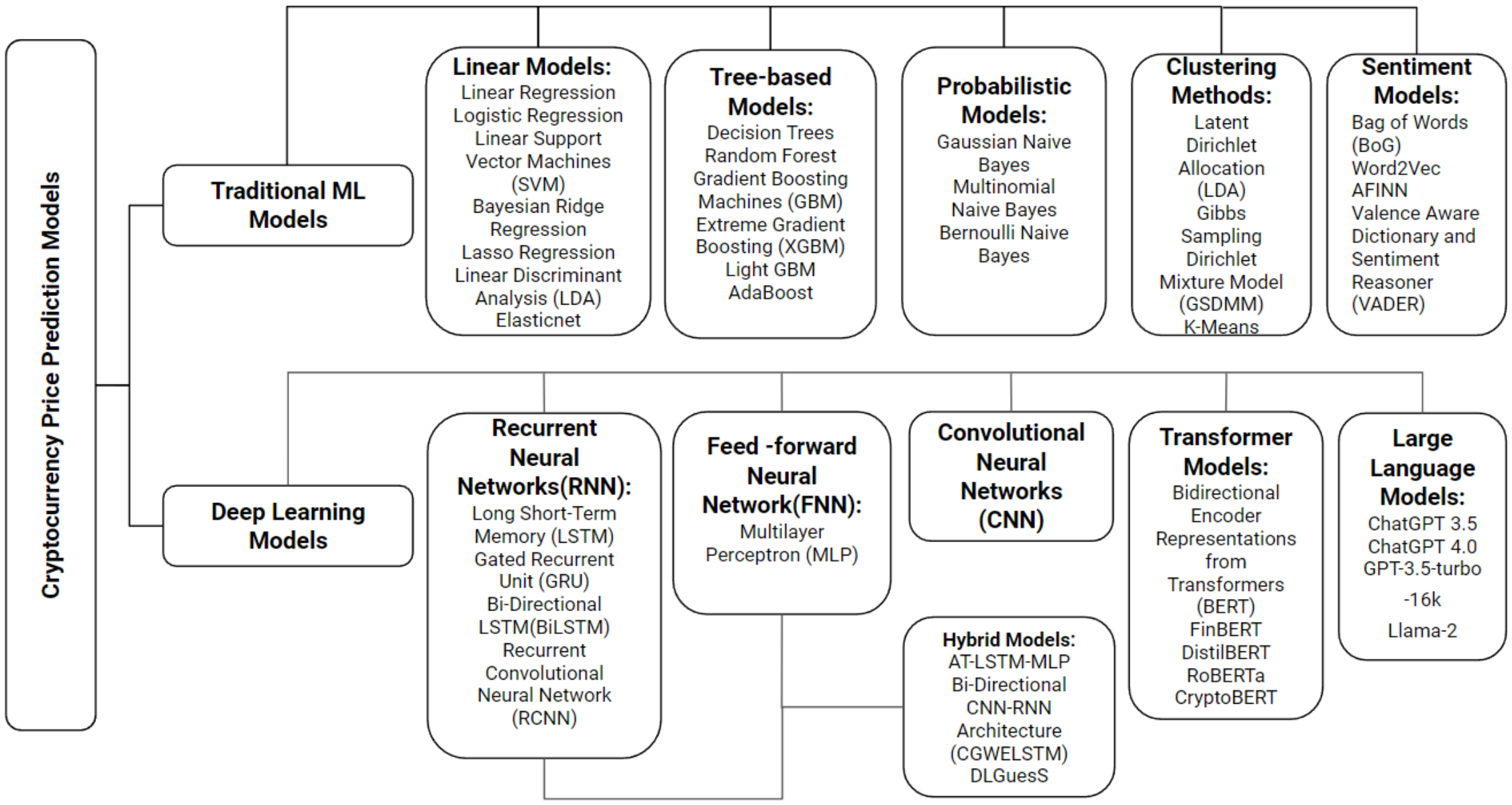

Figure 1: Overview of the Cryptocurrency Price Prediction Models
Source: Jannatun Nayeem Pinky

# 2 Background

The intersection of financial technology (Fintech), cryptocurrencies, and industrial engineering has captured substantial interest in academic literature, highlighting the evolving nature of these fields and their influence on the global financial landscape. With their expertise in process optimization, risk mitigation, and system design, industrial engineers have a unique role in shaping the development and application of advanced technologies within these sectors. In this section, we review key works that examine several aspects of Fintech and cryptocurrencies, organizing them into relevant subthemes for a thorough understanding and identifying where industrial engineers can contribute.

Several studies have sought to define and trace the evolution of Fintech, acknowledging it as a disruptive force within the financial sector (Giglio 2021; Milian, Spinola, and Carvalho 2019; Siddiqui and Rivera 2022). Using systematic literature reviews (SLRs) and qualitative content analysis (QCA), researchers have detailed the attributes, business models, and competitive advantages of Fintech, highlighting its transformative potential and associated challenges (Milian, Spinola, and Carvalho 2019; Siddiqui and Rivera 2022). Bibliometric analyses have mapped Fintech literature's progression and critical aspects, pinpointing significant contributors, research streams, and emerging trends (Bajwa et al. 2022; Wang et al. 2022). These studies provide insights into the global distribution of Fintech research, interdisciplinary collaborations, and citation patterns, aiding in a comprehensive understanding of the field's development and future directions. Industrial engineers can contribute by applying their skills in optimizing complex systems, improving the efficiency of Fintech applications, and addressing scalability and operational challenges.

Research on Fintech applications covers various sectors, such as banking, payments, insurance, wealth management, and regulatory compliance (Fosso Wamba et al. 2020; Siddiqui and Rivera 2022). Case studies and systematic reviews explore the advantages, challenges, and adoption trends of Fintech technologies, offering valuable insights for both practitioners and policymakers. With their background in systems engineering and process management, industrial engineers can play a crucial role in enhancing these Fintech solutions by streamlining workflows, improving risk management frameworks, and facilitating the integration of recent technologies.

Scholarly research on cryptocurrency price prediction has examined traditional statistical methods and machine learning techniques to evaluate their effectiveness in forecasting the highly volatile cryptocurrency markets (Sibel Kervancı and Akay 2020; Khedr et al. 2021). These studies address the difficulties in modeling cryptocurrency prices due to their extreme volatility, limited historical data, and lack of seasonal patterns, stressing the need for diverse datasets and features to enhance prediction accuracy. Here, industrial engineers can contribute by refining the data collection and modeling processes, applying optimization techniques to improve the efficiency and accuracy of these predictive models, and incorporating operational constraints into the forecasting systems.

The behavioral dimensions of cryptocurrencies have also been explored, with researchers investigating herding behavior, sentiment analysis, and investor decision-making in cryptocurrency markets (Almeida and Gonçalves 2023; Ballis and Verousis 2022). Empirical research has uncovered trends in investor behavior, the impact of social factors on investment choices, and the occurrence of speculative bubbles, providing valuable insights for regulators and market participants. Through their knowledge of human factors and decision-making processes, industrial engineers can contribute to designing systems that incorporate behavioral insights to optimize trading strategies and manage market risks.

Additionally, research on cryptocurrency market microstructure has focused on liquidity dynamics, volatility spillovers, and how news sentiment affects price clustering (Almeida and Gonçalves 2024). Systematic reviews have also examined emerging cyber threats to cryptocurrencies and the defensive strategies developed to address these threats, underscoring the necessity for robust cybersecurity measures to protect digital assets (Badawi and Jourdan 2020; Taylor et al. 2020). Industrial engineers can aid in designing resilient security systems by applying systems reliability, risk analysis, and process optimization principles to enhance the cybersecurity infrastructure within cryptocurrency markets.

This chapter highlights a gap in using advanced modeling techniques, such as deep learning algorithms and large language models, for predictive analytics in Fintech and cryptocurrency markets. Additionally, there is a lack of research on combining various modeling approaches and assessing their effectiveness in tackling the challenges of forecasting volatile markets and analyzing investor sentiment. The integration of sentiment analysis with price prediction models remains remarkably underexplored. While some studies have investigated the effect of sentiment on market dynamics, comprehensive models that merge sentiment data with price data for more accurate predictions are still lacking.

To address this gap, the present chapter will thoroughly investigate advanced modeling techniques for predictive analytics in Fintech and cryptocurrency markets, including linear models, tree-based models, probabilistic and clustering models, deep learning algorithms, and large language models (LLMs). Industrial engineers can contribute by optimizing the performance of these models, improving their scalability, and applying process management techniques to streamline the integration of various modeling approaches. The chapter aims to enhance understanding of market dynamics, improve price prediction accuracy, and more effectively analyze investor sentiment. Additionally, it will offer insights into integrating diverse modeling approaches and their impact on decision-making in financial markets, including the combined use of sentiment analysis and price prediction models to capture the comprehensive dynamics of cryptocurrency markets.

# 3 Methods

This section comprehensively explores the theoretical foundations essential for understanding machine learning algorithms. We begin by exploring classical statistical methods to establish a foundational understanding. Subsequent subsections delve into specific models, including clustering methods like Latent Dirichlet Allocation (LDA) and deep learning algorithms such as the Multi-Layer Perceptron (MLP), Long Short-Term Memory (LSTM) networks, Gated Recurrent Unit (GRU), Bidirectional LSTM (BiLSTM), and hybrid architectures like the AT-LSTM-MLP model. Industrial engineers play a pivotal role in optimizing the application of these models by improving data processing workflows, system efficiency, and model performance through advanced operations research techniques. These subsections unravel the intricacies of probability distributions, neural network architectures, and advanced techniques crucial for modern machine learning applications. The insights industrial engineers provide help refine these algorithms and ensure their effective integration into real-world systems, enhancing their capability to address complex challenges in various domains."

## 3.1 Linear Models:

Researchers commonly utilize linear models to analyze the linear relationship between variables. Linear models are preferred for their ease of interpretation and simple operation, making them particularly useful for industrial engineers involved in data-driven decision-making and trend analysis. Moreover, they play a significant role in examining trends in cryptocurrency adoption, providing a straightforward method to model relationships between market data and influencing factors. The use of linear models alone or in comparison to other models is explored in this research, highlighting their practicality for industrial engineers aiming to optimize processes and forecast outcomes in dynamic environments like cryptocurrency markets. Greaves and Au (2015) utilized various machine learning models, including linear regression, logistic regression, support vector machine (SVM), and neural networks, for predicting bitcoin price movements based on transaction data. Linear regression aimed to minimize mean squared error, logistic regression employed maximum likelihood estimation, SVM constructed a separating hyperplane, and neural networks featured a feedforward architecture with dropout to prevent overfitting during training. Industrial engineers, particularly those specializing in financial engineering or operations research, can find these models applicable for enhancing predictive accuracy and streamlining operational efficiency. Abraham et al. (2018) also applied linear regression, logistic regression, and support vector machine models for cryptocurrency price prediction using tweet volumes and sentiment analysis. These models were chosen based on correlation analysis results between input features and price changes. Feature engineering included integrating sentiment analysis scores, Google Trends search volume index (SVI) values, and tweet volume data. This combination of machine learning and social media analytics can be particularly beneficial for industrial engineers working in sectors where consumer behavior and market sentiment impact production forecasts or supply chain optimization.

Similarly, Jain et al. (2018) utilized multiple linear regression (MLR) to model the relationship between tweet sentiment and cryptocurrency prices. Tweet counts tagged with positive, negative, and neutral sentiments served as independent variables, while average price was the dependent variable. The MLR model underwent training and validation, with iterations performed until satisfactory results were achieved. This predictive modeling approach offers industrial engineers insights into how external variables, such as public sentiment, can be incorporated into forecasting models for improved process optimization and resource allocation. Wolk (2020) employed the Valence Aware Dictionary and sentiment Reasoner (VADER) sentiment analysis tool to analyze text data, considering nuances such as punctuation, capitalization, negation, and lexicon values amplification. Various predictive and descriptive models were utilized, including least square linear regression (LSLR), Bayesian ridge regression, AdaBoost, gradient boosting, support vector regression, stochastic gradient descent, multilayer perceptron neural network, decision tree, ElasticNet, and a hybrid approach. These models were implemented using the SKLEARN Python library.

For industrial engineers, these models offer diverse options for solving optimization problems, from process efficiency improvements to complex system simulations in production environments. Wu (2023) employed three machine learning models to predict daily cryptocurrency returns: Ordinary Least Squares Regression, Random Forest Regression, and LSTM neural network. These models were chosen for their suitability in time series modeling and forecasting. Input variables for prediction included lagged daily price returns of cryptocurrencies and average sentiment scores related to Bitcoin; both lagged by one day. A time lag of 60 days (about 2 months) was applied, and the dataset was divided into 80% training and 20% testing sets to evaluate model performance. This methodological approach can guide industrial engineers in adopting time series forecasting models to optimize manufacturing, organization, or financial decision-making operations. McCoy and Rahimi (2020) employed support vector machines (SVM), k-means clustering, and linear regression for predicting cryptocurrency price movements using social media data. Features generated from tweets and trading data were optimized using a genetic algorithm, with hyperparameters tuned to optimize model performance. Models were trained separately for each cryptocurrency using historical data. Industrial engineers can adapt such models for predictive maintenance, resource allocation, or production planning, leveraging optimization techniques to enhance model accuracy.

Furthermore, Zhu et al. (2023) integrated various forecasting algorithms such as support vector regression (SVR), least-squares support vector regression (LSSVR), and twin support vector regression (TWSVR) into their prediction model architecture. These models were chosen for their generalization ability, high prediction accuracy, and robustness to outliers. Optimization algorithms like the Whale Optimization Algorithm (WOA) and Particle Swarm Optimization (PSO) were employed to optimize model parameters. As seen in this study, the integration of optimization algorithms is a valuable tool for industrial engineers aiming to solve complex optimization problems in areas such as supply chain management, production scheduling, and operations research.

### 3.1.1 Linear Regression:

Researchers have explored various methodologies in predictive modeling to elucidate relationships between variables and predicted outcomes. Among these techniques, Least Squares Linear Regression (LSLR) is a commonly employed approach due to its simplicity and interpretability. Industrial engineers, leveraging their expertise in optimizing processes and system efficiency, often utilize LSLR for its ability to provide clear insights into variable relationships. As articulated by Yasir et al. (2023), linear models are favored for their straightforward nature. In the study conducted by Wolk (2020), LSLR was applied using the Python library SKLEARN. LSLR operates on minimizing errors by representing data as a linear combination of independent variables and coefficient matrices. The relationship between the predicted value $\hat{Y}$ and the array of inputs $X$ is expressed as:

$$\hat{Y} = \widehat{\beta_0} + \sum_{j=1}^{p} \widehat{X_J}\widehat{\beta_J} \tag{1}$$

To derive the coefficient matrix $\beta$, the following formula is employed:

$$\beta = X(X^T X)^{-1} X^T Y \tag{2}$$

This formulation encapsulates the essence of LSLR as applied in the study by Wolk (2020), shedding light on how the coefficient matrix is computed to ascertain the predictive relationship between variables. Industrial engineers contribute by optimizing these linear regression models, ensuring that data processing and model implementation are efficient and effective, thus enhancing the accuracy and reliability of predictions in complex systems.

### 3.1.2 Bayesian Ridge Regression:

In predictive modeling, Bayesian ridge regression offers a robust alternative to traditional linear regression techniques by incorporating probabilistic principles. As explored by Wolk (2020), this method introduces a regularization parameter, lambda, which penalizes beta coefficients, effectively guiding them toward zero and preventing overfitting. With their expertise in optimizing systems and managing trade-offs, industrial engineers can effectively leverage Bayesian ridge regression to balance model complexity and accuracy.

According to the study, Bayesian ridge regression yields a probabilistic model characterized by a Gaussian parameter:

$$p(\lambda) = N(\alpha\lambda^{-1}I_p) \qquad (3)$$

The selection of $\alpha$ and $\lambda$ is pivotal in Bayesian ridge regression and is often determined by the gamma distribution. Default values for $\alpha$ and $\lambda$ are typically set to $10^{-6}$, although these parameters can be adjusted using the SKLEARN package to accommodate specific modeling requirements. Industrial engineers can optimize these parameters to enhance model performance, ensuring that it meets the specific needs of complex systems and datasets.

In Bayesian ridge regression, coefficient values are assigned utilizing the following equation:

$$\beta = X(X^T X + \lambda I)^{-1} X^T Y \qquad (4)$$

This formulation encapsulates the essence of Bayesian ridge regression as applied in the study by Wolk (2020), elucidating how coefficient values are computed to balance model complexity and predictive accuracy. Industrial engineers can further refine Bayesian ridge regression models to enhance their effectiveness in real-world applications by incorporating their process optimization and system design skills.

### 3.1.3 Support Vector Machine:

In the realm of predictive modeling, Support Vector Machines (SVM) emerge as a sophisticated supervised learning methodology, adeptly employed for both classification and regression tasks (Jung et al. 2023; Lamon, Nielsen, and Redondo 2017; Ortu et al. 2022; Pant et al. 2018; Valencia, Gómez-Espinosa, and Valdés-Aguirre 2019). Using a kernel function, SVMs seek to delineate hyperplanes within high-dimensional or potentially infinite spaces. With their expertise in optimizing systems and handling complex datasets, industrial engineers can leverage SVMs to improve prediction accuracy, particularly in scenarios with limited or noisy data, which are common in asset return prediction tasks (Akyildirim, Goncu, and Sensoy 2021).

This approach aims to maximize the separation distance between the hyperplane and the nearest training instances. According to Yasir et al. (2023), SVMs are also prevalent in forecasting time series data. Valencia, Gómez-Espinosa, and Valdés-Aguirre (2019) articulated the decision function of SVM as:

$$y = sgn(\sum_{i=1}^{n} y_i \alpha_i K(x_i, x) + \rho) \quad (5)$$

Here, $y_i$ signifies the classification label, assuming values of either 1 or -1, with $n$ denoting the count of training vectors, $\alpha_i$ representing a Lagrange multiplier, $K(x_i, x)$ denoting the Kernel function, and $\rho$ representing the intercept for the maximum margin decision boundary. Industrial engineers can enhance SVM models by optimizing these parameters and improving computational efficiency in large-scale datasets.

As stated by Lamon, Nielsen, and Redondo (2017), the optimization problem for finding the optimal margin is formulated as:

$$\min_{\gamma,\omega,b} \frac{1}{2}|\omega|^2 \quad (6)$$

$$\text{s.t. } y^{(i)}\left(\omega^T x^{(i)} + b\right) \geq 1, i = 1, \ldots, m \quad (7)$$

Industrial engineers can apply their knowledge to refine these optimization processes, ensuring that the SVMs deliver accurate and actionable insights for complex prediction tasks. This will enhance their application in financial and other industrial domains.

Several researchers utilized SVM models to forecast the cryptocurrency price fluctuation by analyzing historical data, market sentiment and technical indicators. For instance, Gurrib and Kamalov (2022) focused on predicting Bitcoin price movements using sentiment analysis alongside machine learning techniques like SVM. After preprocessing the data and conducting sentiment analysis using tools like VADER, they construct a forecasting model comprising Linear Discriminant Analysis (LDA) and sentiment analysis. Sentiment analysis used rule-based (VADER) and automatic (ML-based using TextBlob) approaches. The features of the SVM model include close price return, relative open price difference, relative high-low difference, and various sentiment variables. The data is split chronologically into train/test subsets due to its time-series nature, and optimal regularization parameters are determined through moving window cross-validation. Industrial engineers could apply a similar approach to time-series data in production processes, utilizing SVMs to predict equipment failures or process inefficiencies by analyzing variables like operational performance, machine metrics, and sentiment data from operator logs.

On the other hand, Wimalagunaratne and Poravi (2018) developed a predictive model for the global cryptocurrency market by collecting data from social media platforms, specifically Twitter, using its API. They gather tweets about cryptocurrencies, particularly Bitcoin, to gauge public sentiment. The proposed system architecture comprises multiple components, including capturing tweets using the Twitter API, processing them through sentiment analysis, and storing sentiment information, trading information, and historical prices in a database. Machine learning models such as Support Vector Machines (SVM), Neural Networks, and Linear Regression are trained using the preprocessed data to predict cryptocurrency prices. The models consider public perception captured through sentiment analysis, trading information, and historical prices to predict cryptocurrency prices. Similarly, industrial engineers can collect and process real-time data from manufacturing environments, including sensor data and operator feedback, to predict machine performance, downtime, or supply chain disruptions using SVM models integrated into a data pipeline.

Additionally, Mirtaheri et al. (2021) focused on identifying and analyzing cryptocurrency manipulations in social media. They collect data from various sources, including Telegram, Twitter, and cryptocurrency market data, preprocess it, and extract features for binary classification tasks using Random Forest and linear SVM classifiers. Various features were extracted from the data, including economic features, Twitter statistics, graph features, and target price features. These features were used for binary classification tasks: predicting pump attempts and predicting the success of pump attempts. Binary Random Forest classifiers were trained for both functions, along with linear SVM classifiers using the TF-IDF vector representation. Data was split into training and test sets based on timestamps. Model training involved the iterative addition of new data points to the training set, with evaluation based on the area under the receiver operating characteristic curve (ROC-AUC). Different periods were considered for feature extraction and model training based on ROC-AUC scores. By classifying patterns in production data and supply chain communication, industrial engineers could adapt such classification techniques to identify operational bottlenecks or process manipulations, such as in supply chain fraud detection.

Finally, Mahdi et al. (2021) collected data on COVID-19 cases and deaths, gold prices, and cryptocurrency returns, preprocessed it, and integrated it into a unified format suitable for analysis. They employ Support Vector Machine (SVM) algorithms for analysis, applying probability density functions, performing statistical tests like the Jarque-Bera test, and training SVM models using one-to-one and one-to-rest methods. For industrial engineers, such approaches can be utilized to analyze and model complex systems involving multiple factors, such as production rates, environmental conditions, and machine performance, using SVMs to ensure accurate predictions and reliable outcomes in manufacturing processes.

### 3.1.4 Linear Discriminant Analysis (LDA)

In the landscape of classification modeling, Linear Discriminant Analysis (LDA) stands out as a prominent linear classifier that leverages Bayes' rule and assumptions about class conditional distributions to predict class labels, as discussed by Gurrib and Kamalov (2022). LDA aims to maximize the posterior probability of the class label given a fixed feature vector $X$. Using their systems optimization and data analysis expertise, industrial engineers can apply LDA to effectively classify and analyze complex datasets, ensuring robust decision-making processes.

A fundamental assumption of LDA is that the conditional density function $f_k(x)$ of $X$ given $y = k$ follows a Gaussian distribution, with the covariance matrix Σ of $f_k(x)$ being uniform across all classes $k$. The conditional density function in LDA is expressed as:

$$f_k(x) = \frac{1}{(2\pi)^{p/2}|\Sigma|^{1/2}} e^{\left(-\frac{1}{2}(x-\mu_k)^T \Sigma^{-1}(x-\mu_k)\right)} \tag{8}$$

Here, $\mu_k$ represents the mean vector for class $k$, and Σ denotes the covariance matrix. The decision function in LDA seeks to identify the class with the maximum posterior probability given the feature vector $X$. It is expressed as:

$$y(x) = \text{argmax}\,[-\frac{1}{2}(x-\mu_k)^T\Sigma^{-1}(x-\mu^k) + log(\pi_k))] \tag{9}$$

Here, $\pi_k$ represents the prior probability of class $k$. According to Gurrib and Kamalov (2022), LDA offers two significant advantages: it operates as a linear model, reduces susceptibility to overfitting, and provides a closed-form solution, facilitating the analytical calculation of the decision boundary. Industrial engineers can leverage these advantages to enhance model performance and interpretability, ensuring that LDA models are well-suited for complex classification tasks in various industrial applications.

### 3.1.5 Multiple Linear Regression

In statistical modeling, Multiple Linear Regression (MLR) is a valuable technique, as elucidated by Jain et al. (2018), for capturing the relationship between a dependent variable and two or more independent variables. MLR is instrumental for industrial engineers who need to model complex relationships within large datasets, enabling them to understand and predict how multiple factors collectively influence an outcome. By fitting a linear equation to the data, MLR effectively encapsulates the intricate interplay between variables, providing insights crucial for decision-making and optimization in various industrial contexts.

The formulation of the MLR equation entails forecasting the dependent variable $Y$ as a linear combination of the independent variables $X_1$, $X_2$, $X_3$,...,$X_n$, each weighted by coefficients $b_1$,$b_2$,$b_3$,...,$b_n$, respectively, and augmented by an intercept term $b_0$:

$$Y = b_0 + b_1X_1 + b_2X_2 + b_3X_3 + \cdots + b_nX_n \quad (10)$$

As highlighted by Jain et al. (2018), MLR relies on several key assumptions, including linearity, absence of multicollinearity, equality of variance, normality, and independence of errors. These assumptions are routinely examined using statistical techniques and tools like the Stats model library. For industrial engineers, MLR offers several advantages, including evaluating the magnitude of the effects exerted by independent variables on the dependent variable, predicting future values, and identifying trends within the dataset, supporting data-driven decisions and process improvements in diverse engineering applications.

### 3.1.6 Logistic Regression

Logistic Regression, initially proposed by Cox (1958), is a cornerstone algorithm in machine learning and a critical tool for industrial engineers involved in classification tasks. As discussed by both Lamon, Nielsen, and Redondo (2017) and Jung et al. (2023), Logistic Regression is renowned for its efficacy in predicting the probability of a categorical outcome by leveraging a linear combination of independent variables. This makes it particularly valuable for industrial engineers who need to model binary outcomes, such as defect detection in manufacturing processes or predictive maintenance for machinery.

According to Akyildirim, Goncu, and Sensoy (2021), Logistic Regression can be conceptualized as a single-layer neural network specifically designed for binary response variables. The sigmoid function encapsulates the logistic hypothesis, denoted as $\sigma(x)$, which transforms the dot product of a weights vector $W$ with the features $x$ into the probability of an example belonging to class 1. The sigmoid function is mathematically defined as:

$$\sigma(x) = \frac{1}{1 + e^{-W^T x}} \quad (11)$$

Furthermore, the logistic loss function is crucial for computing the error associated with Logistic Regression. It is expressed as:

$$l(x) = \sum_{i=1}^{m} y^{(i)} \, log(\sigma(x^{(i)})) + (1 - y^{(i)}) \, log(1 - \sigma(x^{(i)})) \quad (12)$$

Both Lamon, Nielsen, and Redondo (2017) and Jung et al. (2023) underscored the simplicity and effectiveness of Logistic Regression, particularly in binary classification tasks. This technique provides valuable insights into model accuracy and error minimization for industrial engineers, enhancing decision-making processes in applications such as quality control, risk assessment, and operational optimization.

## 3.2 Tree-Based Models

Tree-based models, such as Random forest and Gradient Boosting, have been widely utilized to capture the price movements of cryptocurrency market due to their ability to efficiently analyze patterns in large dataset.The papers reviewed present a variety of approaches utilizing tree-based models for sentiment analysis and price prediction in the cryptocurrency domain. These approaches have significant implications for industrial engineers, especially those focused on decision-making optimization, data analysis, and process automation in dynamic environments. Industrial engineers can leverage these models to improve forecasting accuracy and optimize complex systems, particularly in sectors where vast data inputs and rapid decision cycles are critical. Şaşmaz and Tek (2021) focused on sentiment analysis using Twitter data and daily cryptocurrency prices, employing a Random Forest (RF) model. The RF model was optimized using GridSearchCV to tune hyperparameters such as the number of estimators, max features, max depth, and criterion. Input data were converted into token counts using CountVectorizer and transformed using TF-IDF.For industrial engineers, using tree-based models like RF, with techniques such as hyperparameter tuning, offers an opportunity to optimize production processes, improve quality control, and enhance predictive maintenance models by identifying patterns in vast amounts of data. Inamdar et al. (2019) utilized the Random Forest Regression algorithm for predicting cryptocurrency values. The RandomForestRegressor class from the sklearn.ensemble package was employed with bootstrapping enabled to prevent overfitting. Model accuracy was assessed during training using the out-of-bag (OOB) score parameter, and model parameters such as estimator attributes were adjusted for optimization. This methodology can be useful for industrial engineers when analyzing system performance or optimizing processes where historical data and real-time feedback improve efficiency.

Similarly, Rahman et al. (2018) collected Twitter data for sentiment analysis, preprocessing the text to clean irrelevant elements and calculate sentiment scores. They prepared two datasets based on machine learning algorithm requirements: one with a single sentiment score and another with sentiment scores divided into ten emotions. Various regression and classification models, including Random Forest Regression, were implemented using these datasets. For industrial engineers, such models are vital for sentiment analysis in manufacturing environments or for gauging consumer sentiment, which can directly influence product development cycles and market demand forecasting. Fareed et al. (2024) conducted emotion detection in cryptocurrency-related tweets using machine learning algorithms. They employed supervised classification models, including Random Forest, Decision Tree, AdaBoost Classifier, Stochastic Gradient Descent Classifier, Extra Tree Classifier, and Gradient Boosting Classifier, implemented using libraries such as Scikit-learn and NLTK. The training dataset was split into training and testing sets using an 80-20 ratio, and each model was fine-tuned with specific hyperparameters to optimize performance. Industrial engineers can adapt these models for classification tasks in supply chain management, quality assurance, or workforce optimization, allowing for better real-time decision-making in operations. Jung et al. (2023) focused on predicting Bitcoin trends using machine learning and sentiment analysis alongside technical indicators. They collected a comprehensive dataset, including news articles, Reddit submissions/comments, and Bitcoin price data. After preprocessing, which involved calculating daily price fluctuations and computing technical indicators, sentiment analysis was performed using VADER. Six classifiers were employed: Logistic Regression, Naıve Bayes, Support Vector Machine (SVM), Random Forest, XGBoost, and LightGBM. Feature scaling was applied using MinMaxScaler, and hyperparameter tuning was conducted using grid search with 5-fold cross-validation. This meticulous approach ensures robust model training and evaluation, offering industrial engineers insights into optimizing predictive models for production schedules, inventory management, or demand forecasting based on technical indicators and external data sources.

Lastly, Basher and Sadorsky (2022) investigated the impact of macroeconomic news on cryptocurrency returns, utilizing tree-based methods such as random forests and bagging. Various prediction models, including random forests and tree bagging, were employed with different configurations, such as the number of trees and randomly chosen predictors at each split. Tuned random forest models were also utilized, where the number of predictors was determined through cross-validation. Boosting was implemented by fitting successive models to residuals, with the number of iterations determined via cross-validation. Industrial engineers can use similar techniques, like boosting and bagging, to improve system reliability, optimize process outcomes, or forecast future trends in large-scale industrial systems by learning from past performance and external factors.

### 3.2.1 Random Forest

Random Forest stands out as a pivotal ensemble learning algorithm, harnessed for classification tasks (Jung et al. 2023; Dudek et al. 2024; and Pant et al. 2018). This method operates by constructing multiple decision trees during the training phase. Subsequently, when predicting the final output class, the algorithm aggregates the predictions from each tree and selects the mode that represents the most commonly occurring class among all the predictions. Industrial engineers often leverage Random Forest algorithms to enhance the robustness of predictive models in complex systems, particularly in domains where reliable decision-making is critical. For example, Random Forest can optimize operational efficiency in financial sectors, including cryptocurrency systems, by enhancing the detection of anomalies and mitigating risks of overfitting. According to Akyildirim, Goncu, and Sensoy (2021), the primary benefit of employing the Random Forest algorithm lies in its ability to mitigate the risk of overfitting and minimize the duration of training. Mathematically, the ensemble of decision trees $R$ is represented by Pant et al. (2018) as:

$$R = [h(x|\phi 1), \ldots\ldots\ldots\ldots, h(x|\phi k)] \tag{13}$$

The final predicted class $Y$ is determined by computing the mode of the predicted classes from each tree, denoted as:

$$Y = MODE[h(y_1) \ldots\ldots, h(y_k)] \tag{14}$$

Here, $h(x|\phi)$ represents a decision tree with parameters $\phi$, $h(y)$ is the predicted class of the decision tree, $R$ is the ensemble of decision trees, and $Y$ is the class with the maximum votes.

### 3.2.2 Gradient Boosting

Boosting algorithms like AdaBoost and Gradient Boosting operate iteratively to minimize errors and enhance model performance (Wolk 2020). These methods are precious in engineering applications, where precision and efficiency are critical. Industrial engineers often utilize gradient-boosting techniques to optimize complex systems, including production processes and supply chains, by improving predictive accuracy and decision-making. The iterative nature of these algorithms is encapsulated in the equation:

$$E_{\mathrm{t}} = \sum_{i} E\,(f_{\mathrm{t-1}}(X_{\mathrm{i}}) + \alpha_{\mathrm{t}}\, h(x_{\mathrm{i}})) \tag{15}$$

Here, $E$ denotes the error, $f_{t-1}(X_i)$ represents the previous model's prediction, and $\alpha_t h(x_i)$ signifies the contribution of the weak learner, weighted accordingly.

In Gradient Boosting, model updates occur through the steepest descent, where the model iteratively minimizes the loss function $L$ by computing the derivative of the residuals and a multiplier:

$$r = \frac{dL(y_i, F(X_i))}{dF(X_i)} \tag{16}$$

$$m = argmin \sum_{i=1}^{n} L\left(y_i, F_{m-1}(X_i) + m\, h_m(X_i)\right) \tag{17}$$

By applying gradient boosting, industrial engineers can optimize the performance of predictive models in areas such as quality control, risk assessment, and process optimization, enhancing overall operational efficiency.

### 3.2.3 XGBoost:

XGBoost is a potent ensemble technique renowned for amalgamating multiple weak decision trees into a robust prediction model (Jung et al. 2023 and Ortu et al. 2022). Due to its capacity for handling large-scale datasets and reducing prediction errors, industrial engineers often leverage XGBoost to optimize complex manufacturing systems, supply chain processes, and operational efficiencies. This methodology involves assigning weights to the learning errors of weak models and iteratively refining predictions in subsequent models. Despite its notable advantage of high-speed learning and classification facilitated by parallel processing, XGBoost is susceptible to overfitting if model parameters are not appropriately tuned. The primary objective of XGBoost is to minimize the following objective function:

$$L(\phi) = \sum_{i} l\,(\hat{y}_i, y_i) + \sum_{k} \omega\,(f_k) \tag{18}$$

In the above equation, $l$ represents the loss function quantifying the discrepancy between the predicted $\hat{y}_i$ and the true $y_i$, while $\omega$ is a regularization parameter. By fine-tuning these parameters, industrial engineers can optimize predictive models for various applications such as demand forecasting, inventory management, and process optimization.

### 3.2.4 LightGBM

LightGBM represents a decision-tree-based learning algorithm within a gradient-boosting framework (Jung et al. 2023). Due to its ability to process large-scale datasets efficiently, industrial engineers frequently utilize LightGBM in operational optimization scenarios such as production planning, resource allocation, and predictive maintenance. Unlike traditional decision tree algorithms that expand trees level-wise, LightGBM adopts a leaf-wise expansion approach. This methodology entails continuously splitting leaf nodes based on maximum delta loss, resulting in deep and asymmetrical trees.

By prioritizing continuous leaf node division over tree balancing, LightGBM aims to minimize prediction error more effectively than balanced tree division methods. This unique feature enables LightGBM to handle large datasets while consuming less memory during execution efficiently. However, for industrial engineers aiming to optimize processes like demand forecasting or quality control, optimal performance with LightGBM is contingent upon appropriate parameter tuning.

## 3.3 Probabilistic and Clustering Models

Probabilistic and clustering models are widely used in the cryptocurrency domain for sentiment analysis and topic modeling, providing valuable insights into the behavior of market participants. These models are applicable in the financial domain and can also be of significant value to industrial engineers. By analyzing patterns and trends within large datasets, industrial engineers can leverage such models for process optimization, fault detection, and system improvement in complex industrial systems. McAteer (2014) used the Naïve Bayes algorithm, a simple probabilistic classifier, for sentiment analysis. It treats tweets as a ”bag of words” and calculates conditional probabilities to categorize tweets as positive, negative, or neutral. This method provides a simple yet effective way to gauge sentiment. For industrial engineers, Naïve Bayes could be used to classify maintenance reports, customer feedback, or machine logs, helping them quickly identify critical issues or areas for improvement in manufacturing processes. Lamon, Nielsen, and Redondo (2017) conducted cryptocurrency price prediction using news headlines and social media sentiment. They preprocess headlines and tweets, extract features using the Count Vectorizer method, and employ classification models like Multinomial Naive Bayes and Bernoulli Naive Bayes. Naïve Bayes assumes conditional independence of features given labels and uses maximum likelihood estimation. The dataset was divided into train (60%), development (20%), and test sets (20%) based on different date ranges.

Industrial engineers could apply similar techniques to classify production data, customer reviews, or supply chain communications, allowing them to make more informed decisions on production schedules, quality control, and supply chain optimization. Coulter (2022) employed a comprehensive preprocessing pipeline using Python libraries such as SpaCy, Gensim, and Pandas. This included steps like parsing, tagging, lemmatization, and tokenization of the text data. Subsequently, they trained a Latent Dirichlet Allocation (LDA) model to identify topics within the corpus of news articles. The LDA model was trained with default hyperparameters, such as alpha and beta values. The number of topics was determined based on coherence scores, with 18 topics identified as optimal. Coherence scores were calculated for different numbers of topics to select the most coherent model. The LDA model was then trained iteratively with different numbers of topics to identify the configuration with the highest coherence score. Once the optimal configuration was determined, the LDA model was trained using the selected number of topics and default hyperparameters. For industrial engineers, LDA models can be used for clustering and topic modeling in various areas, such as analyzing operational reports or maintenance logs, helping to identify recurring issues, patterns, or optimization opportunities in manufacturing systems.

Similarly, Ni, Härdle, and Xie (2020) utilized Latent Dirichlet Allocation (LDA) for topic modeling to identify policy-related news articles and calculate the Hellinger distance to measure similarity. Articles close to the average distance of policy-related news were classified as policy-related, and a threshold was set using the quantile of the average distances for classification. Additionally, they constructed the Cryptocurrency Regulatory News Index (CRRIX) based on the coverage frequency of policy-related news. For industrial engineers, the ability to classify articles or data based on similarity could be applied to process data, product reviews, or defect analysis, helping to optimize systems, reduce production errors, and enhance overall process efficiency.

### 3.3.1 Naive Bayes

Naive Bayes, as discussed by both Lamon, Nielsen, and Redondo (2017) and Pant et al. (2018), is a classification algorithm based on Bayesian probability theory. It assumes conditional independence among features given the class labels, enabling it to efficiently model the joint likelihood of the data. For industrial engineers optimizing cryptocurrency systems, Naive Bayes provides a fast and efficient way to classify market trends or detect security threats by leveraging historical data. Mathematically, Naive Bayes utilizes maximum likelihood estimation to maximize the joint likelihood of the data, which is expressed by Lamon, Nielsen, and Redondo (2017) as:

$$L(X,Y) = \prod_{i=1}^{m} p\,(x^{(i)}, y^{(i)}) \tag{19}$$

Also, Naive Bayes assumes feature independence (Wong 2021 and Pant et al. 2018). This assumption allows Naive Bayes to calculate the conditional probability of a class given a feature vector using Bayes' theorem:

$$p(C_k \mid x) = \frac{p(C_k)\, p(x \mid C_k)}{p(x)} \tag{20}$$

Here, $p(C_k)$ represents the prior probability of class $C$, $p(x|C_k)$ represents the class-conditional feature probability, and $p(C_k|x)$ represents the probability of $x$ belonging to class $C_k$. For industrial engineers working with cryptocurrency systems, this probabilistic framework enables them to make informed decisions in fraud detection, price prediction, and risk management. Various variations of Naive Bayes, such as Multinomial and Bernoulli, are commonly utilized in practice, offering flexibility in modeling several types of data distributions.

### 3.3.2 Latent Dirichlet Allocation (LDA)

According to Blei (2012), Latent Dirichlet Allocation (LDA) is an unsupervised machine learning approach that discerns latent topics within a corpus, such as financial news or cryptocurrency market reports. For industrial engineers optimizing cryptocurrency systems, LDA is particularly valuable in uncovering hidden patterns within unstructured data, such as news articles or social media posts influencing market movements. By analyzing large datasets, LDA extracts thematic structures, which can be leveraged for predictive modeling and anomaly detection, crucial in identifying emerging trends or risks in cryptocurrency markets. Ni, Härdle, and Xie (2020) notes that LDA operates through a generative statistical approach, identifying word distributions contributing to topics while simultaneously assigning probabilities of topics to documents. This process assigns specific topics (e.g., market volatility or regulatory changes) to collections of cryptocurrency-related text, enabling engineers to cluster and analyze market sentiment.

In this model, each topic ($z$) is associated with a collection of the most probable words ($w$), and each document ($d$) is annotated with its most probable topics ($z$), making it a robust tool for categorizing news, reports, and communications affecting cryptocurrency trends. Industrial engineers can utilize the probability distributions $p(w|z)$ and $p(z|d)$ to enhance decision-making in algorithmic trading and risk management systems. This can provide insights into how specific external factors, such as geopolitical events or technological announcements, may impact market dynamics.

For effective implementation, industrial engineers can employ Python packages like Gensim, which optimize LDA parameters ($\alpha$ and $\beta$), improving topic coherence and document classification. This enables the development of cryptocurrency systems that forecast market fluctuations based on thematic shifts in relevant external information.

## 3.4 Deep Learning Algorithms

This section explores various neural network models for predicting cryptocurrency price movements through sentiment analysis and historical data. The approaches vary across architectures, data sources, and model types, reflecting the diversity in cryptocurrency forecasting methods.

In Valencia, Gómez-Espinosa, and Valdés-Aguirre(2019), a combination of machine learning models like Multi-Layer Perceptrons (MLPs), Support Vector Machines (SVMs), and Random Forests (RFs) are used to predict cryptocurrency price changes. The feature vectors include market data, sentiment metrics, and polarization scores, with target variables representing binary price movements. Vo, Nguyen, and Ock (2019) used Long Short-Term Memory (LSTM) networks for cryptocurrency price prediction, combining historical price data with semantic vectors from news articles for time series prediction. The LSTM network captures sentiment and price data patterns over time, enhancing prediction accuracy. Raju and Tarif (2020) utilized LSTM networks to predict real-time Bitcoin prices, integrating Twitter sentiment and market data. The architecture is trained using Root Mean Squared Error (RMSE) and compared to an ARIMA model for performance. Habek, Toçoğlu, and Onan (2022) develop a complex CGWELSTM architecture with several layers, including convolutional, bidirectional LSTM, and attention mechanisms. Their model is fine-tuned with hyperparameters to enhance the prediction accuracy for Turkish cryptocurrency-related sentiment and price data. A study by Huang et al. (2021) explores LSTM models for sentiment analysis based on user posts from Sina-Weibo, leveraging tokenized crypto word embeddings. The LSTM model's output is used for price movement prediction based on sentiment. Inamdar et al. (2019) employ LSTM units for sentiment analysis of Twitter and news data, followed by Random Forest Regression for price prediction, highlighting the versatility of combining RNNs with traditional machine learning methods.

The DL-GuesS model by Parekh et al. (2022) integrates LSTM and GRU layers for multicryptocurrency prediction (Bitcoin, Litecoin, Dash), with each asset's sentiment and price history contributing to the final prediction. Pant et al. (2018) investigate Bitcoin price prediction using LSTM and GRU-based RNNs, extracting features using Word2Vec and Bag-of-Words methods for sentiment analysis. This hybrid approach enhances prediction accuracy.

A graph-based sentiment analysis method is proposed in Moudhich and Fennan (2024), where DeepWalk graph embedding techniques generate features for a Bi-LSTM model. This innovative approach aims to improve understanding of relationships within text data for cryptocurrency predictions. In Poongodi et al. (2021), a global cryptocurrency trend prediction system is developed using LSTM layers and StarSpace word embeddings—the model also experimented with various labeling and normalization techniques for Bitcoin price prediction. The work by Ortu et al. (2022) introduces deep learning models like MLPs, LSTMs, and CNNs for price classification based on technical and social media indicators, with hyperparameters optimized using Grid Search. Cheng et al. (2024) combined LSTM, SARIMA, and Facebook Prophet models for forecasting Bitcoin prices, emphasizing the ability of LSTM to capture long-term dependencies in time series data, compared to more traditional methods like SARIMA. A business intelligence model in Yasir et al. (2023) integrated social media sentiment into linear regression, SVR, and deep learning models for cryptocurrency forecasting, focusing on transaction volumes, market capitalization, and social media sentiment.

The NLP and time series forecasting integration seen in Bao (2022) combined sentiment analysis using neural networks and time series forecasting with tspDB, enhancing the precision of predictions based on large datasets. Wong (2021) utilized NaïveBayes and LSTM models for predicting cryptocurrency prices, while Onyekwere, Ogwueleka, and Irhebhude (2022) focused on different MLP designs for optimizing training performance with early stopping mechanisms. Lastly, hybrid models like those in Zahid, Iqbal, and Koutmos (2022) leveraged GARCH, LSTM, and BiLSTM for predicting Bitcoin volatility, combining statistical methods with deep learning architectures to better capture the complexity of cryptocurrency price movements.

These studies demonstrate the evolving landscape of neural network models and their applications in cryptocurrency forecasting. They integrate various data sources, model types, and techniques to improve prediction accuracy.

### 3.4.1 Multi-Layer Perceptron

The Multi-Layer Perceptron (MLP) represents a sophisticated advancement over the traditional perceptron model, addressing its limitations by integrating multiple layers of neurons (Onyekwere, Ogwueleka, and Irhebhude 2022). As highlighted by Dudek et al. (2024), MLPs are extensively employed in regression and classification tasks within industrial engineering domains due to their robust features, including universal approximation capabilities and proficiency in modeling nonlinear relationships. These networks consist of at least three layers: an input layer, one or more hidden layers, and an output layer. The MLP's ability to map input features to target outputs is mathematically represented as:

$$F() : R_{\mathrm{m}} \rightarrow R_{\mathrm{o}} \tag{21}$$

where $m$ signifies the feature vector dimension, and $o$ denotes the target dimension. In industrial applications such as optimizing cryptocurrency systems, MLPs offer significant advantages including high parallel processing efficiency, resilience to noisy data, and the ability to learn complex patterns.

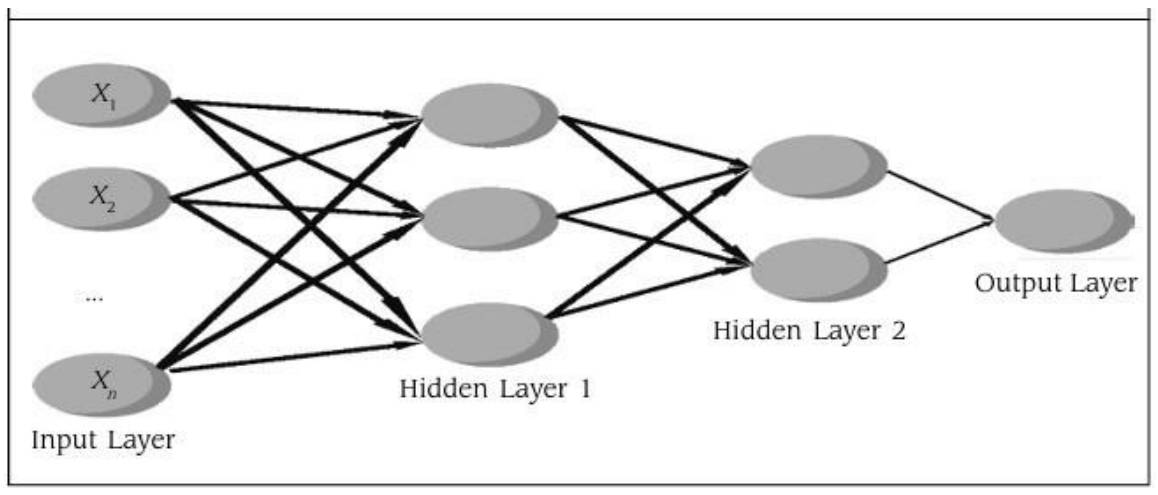

Figure 2: The Architecture of MLP Model
Source: (Onyekwere, Ogwueleka, and Irhebhude 2022)

To prevent overfitting, industrial engineers often use heuristics to determine the optimal number of hidden layers ($N_h$), based on sample size ($N_s$), input neurons ($N_i$), and output neurons ($N_o$). Figure 2 illustrates the architecture of an MLP, demonstrating its layered structure and the role of each layer in transforming input data into predictions.

### 3.4.2 Long Short-Term Memory

Long Short-Term Memory (LSTM) networks, a sophisticated type of recurrent neural network (RNN), are highly effective in modeling and forecasting complex temporal sequences, which is crucial in fields such as industrial engineering. As highlighted by Cheng et al. (2024) and Wong (2021), LSTMs excel in capturing long-term dependencies within sequential data, such as time series of industrial processes, financial market trends, and equipment sensor data. Unlike traditional RNNs, LSTMs incorporate memory cells that preserve information across time steps, allowing for more accurate predictions in scenarios where past data points significantly influence future outcomes. According to Ortu et al. (2022), LSTMs are particularly valuable in optimizing systems like cryptocurrency trading algorithms and predictive maintenance schedules, where understanding historical data trends is critical. As detailed by Dudek et al. (2024), the LSTM architecture includes three primary gates—forget gate, input gate, and output gate—that regulate the flow of information through the network. These gates manage the retention of relevant data and the integration of latest information, ensuring that LSTMs can efficiently handle long-term dependencies. The mathematical expressions governing these gates are crucial for fine-tuning LSTM performance, as shown by the following equations:

**Forget Gate:**

$$f(t) = \sigma(x(t) \cdot U_f + h(t-1) \cdot W_f) \tag{22}$$

**Input Gate:**

$$i_1(t) = \sigma(x(t) * U_i + h(t-1) * W_i \quad (23)$$

$$i_2(t) = \tanh\,(x(t) * U_g + h(t-1) * W_g \quad (24)$$

$$i(t) = i_1(t) * i_2(t) \quad (25)$$

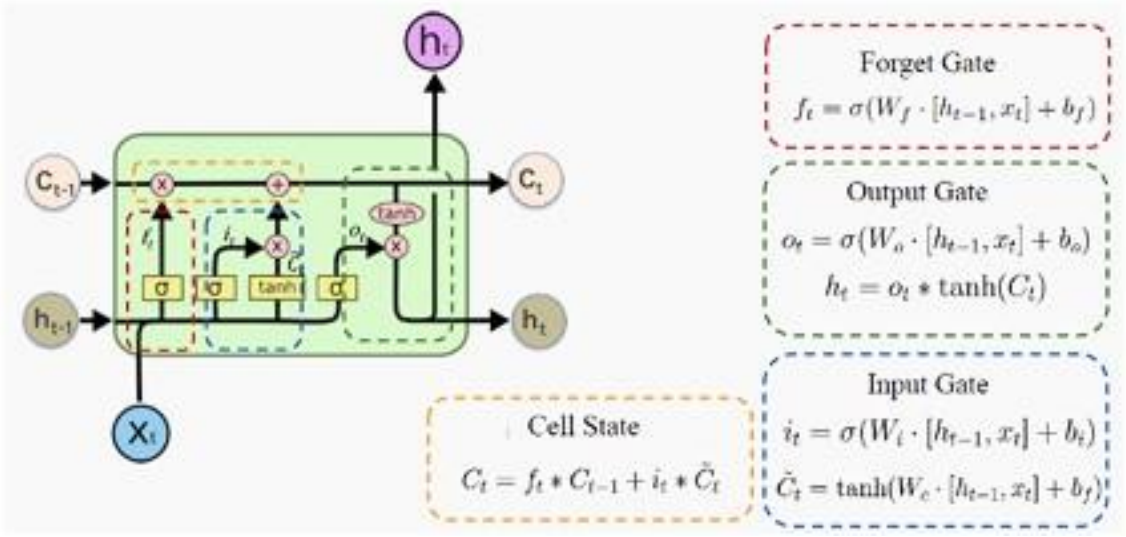

Figure 3: LSTM Model
Source: (Wong 2021)

Figure 3 illustrates the LSTM model, highlighting how these gates operate to maintain an effective memory mechanism, crucial for making reliable forecasts and optimizing systems in industrial engineering applications.

### 3.4.3 Gated Recurrent Unit

Cho(2014) introduced a more recent variation of LSTM, called the Gated Recurrent Unit (GRU), tailored for sequential data analysis. It amalgamates the forget gate and input gate into a single update gate, $z_t$, and merges the cell state $c_t$ and hidden state $h_t$. As stated by Zahid, Iqbal, and Koutmos (2022) the GRU's structure is simpler compared to LSTM, defining the hidden state $h_t$ generated at each time step $t$ through equations (28) to (31).

$$z_t = \sigma(w_z[h_t, x_t]) \quad (27)$$

$$r_t = \sigma(w_z[h_{t-1}, x_t]) \quad (28)$$

$$\widehat{h_t} = \tanh\,(w.[r_t \Delta ht - 1, x_t]) \quad (29)$$

$$h_t = (1 - z_t)\Delta h_{t-1} + z_t \Delta\, \widehat{h_t} \quad (30)$$

Zahid, Iqbal, and Koutmos (2022) also mentioned that the GRU model streamlines LSTM's architectural design by incorporating a single update gate, potentially enhancing its ability to learn sequential data patterns. GRU models entail fewer tensor operations, rendering them faster to train than LSTM. In financial market analysis, researchers often experiment with both LSTM and GRU models to ascertain which one performs better on diverse datasets and market conditions.

### 3.4.4 Bidirectional LSTM

Bidirectional LSTM (BiLSTM) networks, introduced by Graves and Schmidhuber (2005), enhance traditional recurrent neural networks (RNNs) by integrating both forward and backward LSTM architectures. As noted by Zahid, Iqbal, and Koutmos (2022), this dual approach allows BiLSTM to capture long-term dependencies from both past and future observations, providing a more comprehensive understanding of temporal sequences. This is particularly advantageous in industrial engineering applications where data from both directions—historical and future—can significantly impact decision-making and system optimization. For example, in predictive maintenance, BiLSTM can use information from both before and after a failure event to improve predictions and preventative measures. The BiLSTM architecture features forward and backward hidden layers, as depicted in Figure 5, which facilitate the simultaneous processing of data in both temporal directions. This structure allows BiLSTM to enhance its modeling capabilities for complex time series data, such as those found in industrial operations and cryptocurrency market forecasting.

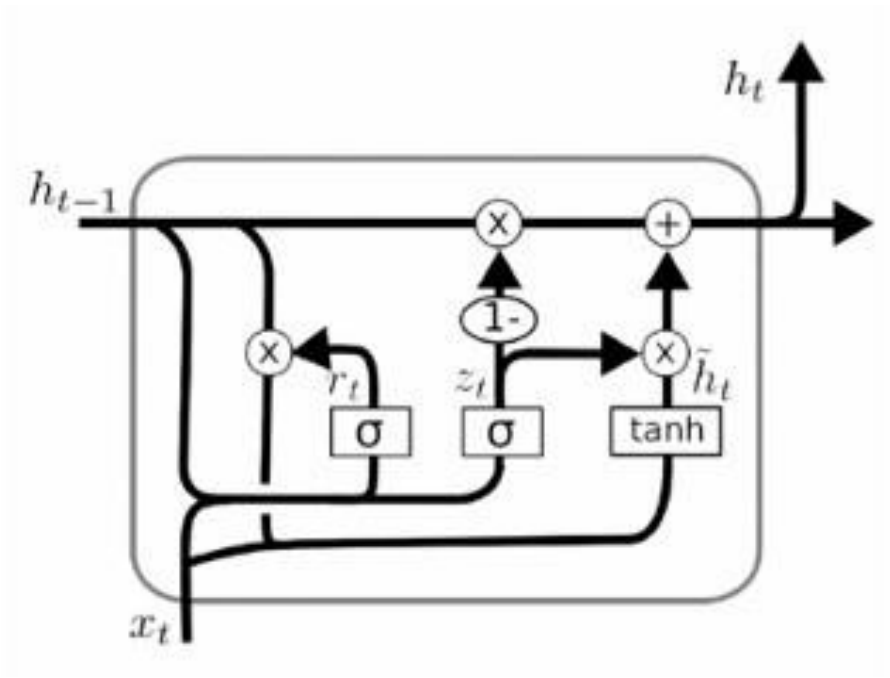

Figure 4: The architecture of the GRU
Source: (Zahid, Iqbal, and Koutmos 2022)

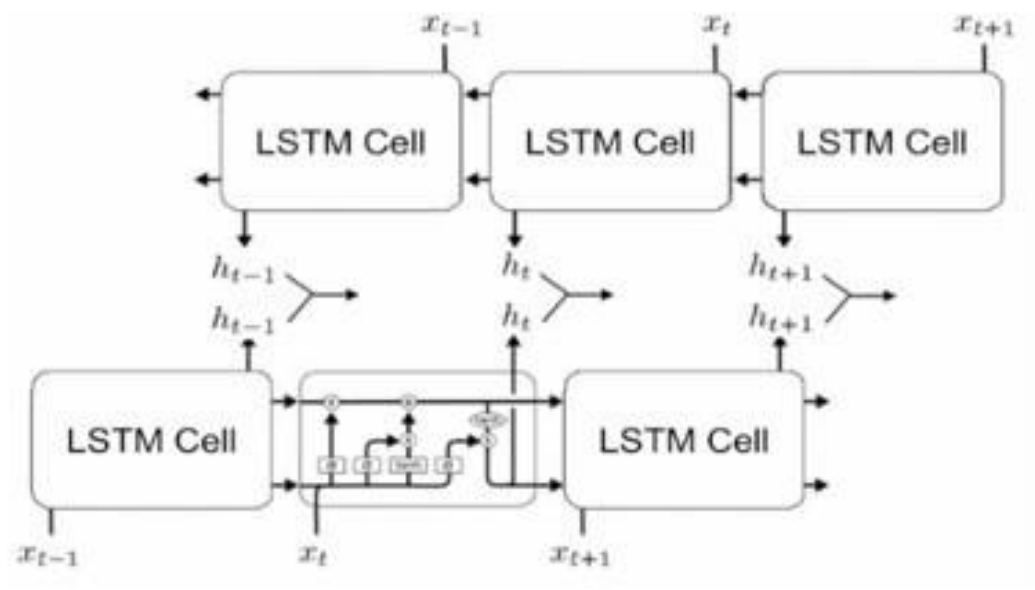

Figure 5: The architecture of BiLSTM Cell
Source: (Zahid, Iqbal, and Koutmos 2022)

### 3.4.5 Bi-Directional CNN-RNN Architecture (CGWELSTM)

In their investigation, Habek, Toçoğlu, and Onan (2022) introduced the CGWELSTM architecture tailored for text sentiment classification, amalgamating convolutional layers and bidirectional Long Short-Term Memory (LSTM) units to extract local and global features from textual data. The architecture comprises seven key modules: input layer, embedding layer, convolution layer, groupwise enhancement mechanism, bidirectional layer, attention mechanism, and fully connected layer, as shown in Figure 6.

**Embedding Layer:** In the Embedding Layer, words within an n-dimensional text document are transformed into V-dimensional word vectors using a pre-trained word2vec embedding matrix, resulting in the word embedding matrix:

$$X = [x_1, x_2, \dots \dots, x_n] \epsilon R^{nxv} \tag{31}$$

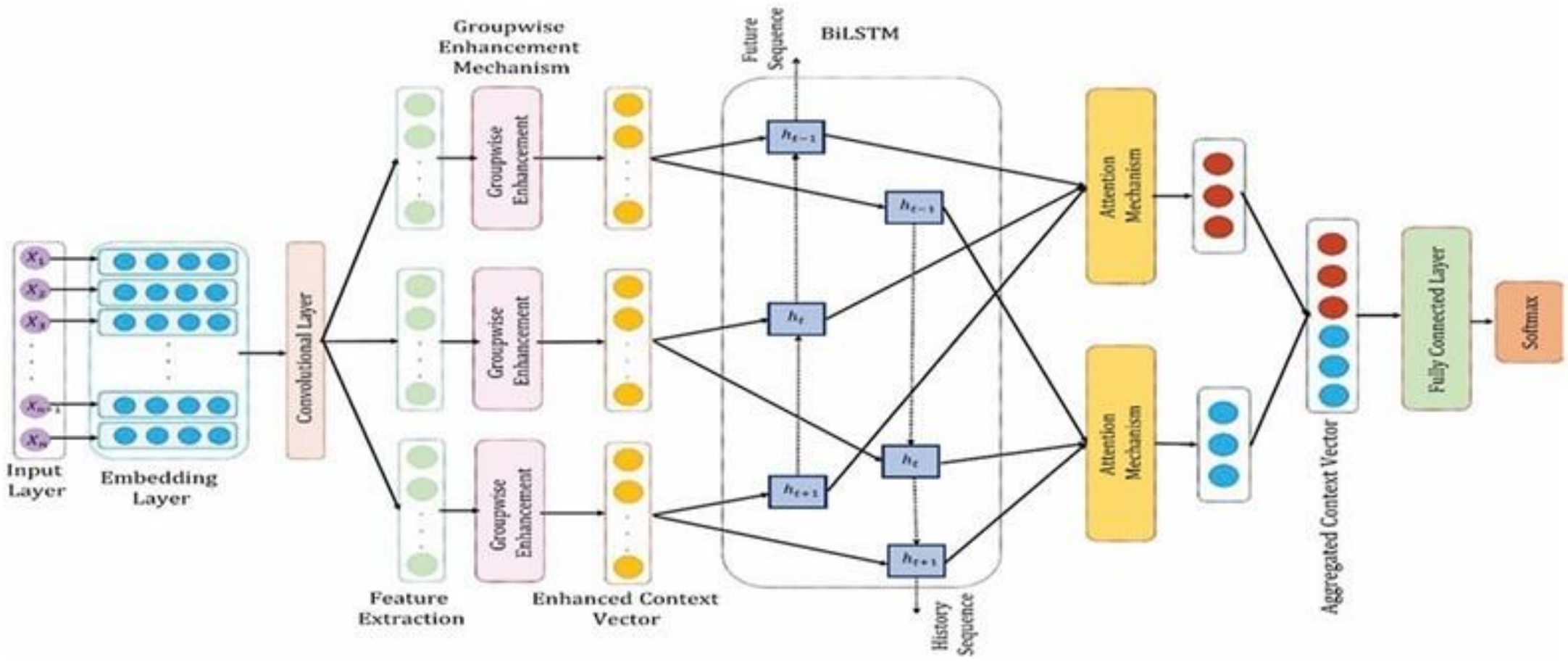

Figure 6: CGWELSTM Model
Source: (Habek, Toçoğlu, and Onan 2022)

**Convolutional Layer:** The convolutional layer applies convolutions on the word embedding matrix $X$ to extract local features. The local context characteristics are derived using the equation:

$$m_i^j = f(x_{i:i+w-1} * F^j + b_0) \tag{32}$$

Where $f$ denotes the RELU, $*$ represents convolution, and w denotes the size of the sliding window.

**Group-Wise Enhancement Mechanism:** The Group-Wise Enhancement Mechanism enhances informative features acquired from the convolutional layer by evaluating each feature's significance based on its similarity with global semantic features. Feature normalization using dot products is employed to prevent biased values, and the improved vector is obtained using parameters $\gamma$ and $\beta$. The equation for feature enhancement is:

$$a_i^k = \gamma c_i^{*k} + \beta \tag{33}$$

where $c_i^{*k}$ is the normalized features and $\gamma$ and $\beta$ are parameters.

**Bidirectional Layer:** The Bidirectional Layer utilizes a bidirectional LSTM to capture features by processing data in both forward and backward directions. The bidirectional LSTM equations include:

$$c_t^{\rightarrow}, h_{tLSTM}^{\rightarrow} = LSTM^{\rightarrow}(c_t^{\rightarrow}, H_{t-1LSTM}^{\rightarrow}, x_t^{\rightarrow}) \tag{34}$$

$$c_t^{\leftarrow}, h_{tLSTM}^{\leftarrow} = LSTM^{\leftarrow}(c_t^{\leftarrow}, h_{t-1LSTM}^{\leftarrow}, x_t^{\leftarrow}) \tag{35}$$

$$h_{tLSTM} = [h_{tLSTM}^{\rightarrow}, h_{tLSTM}^{\leftarrow}] \tag{36}$$

In this context, $h_{tLSTM}$ refers to the hidden state vector, while $c_t$ denotes the cell state vector.

**The Attention Mechanism:** The attention mechanism assigns different weights to features, allowing the model to focus on important parts of the input. The attention mechanism is employed to assign weights to features using the following equations:

$$\alpha_t = \frac{\exp(v^t . u_t)}{\sum_t \exp(v . u_t)} \tag{37}$$

$$S_{A_w} = \sum_t \alpha_t h_t \tag{38}$$

Where $u_t$ and $h_t$ represent the LSTM vectors and $v$ denotes to a trainable parameter. The Fully Connected Layer: The Fully Connected Layer comprises initial layers employing the rectified linear unit (RELU) function and a final layer utilizing the softmax activation function. L2-norm regularization is applied in this layer, with a regularization coefficient of 0.1 and a dropout parameter of 0.30. The binary cross-entropy function serves as the loss function.

In industrial engineering, the CGWELSTM model's ability to effectively capture and process complex, high-dimensional data makes it applicable for optimizing various systems, including predictive maintenance and market forecasting in cryptocurrency systems. Integrating convolutional and bidirectional LSTM layers allows for enhanced feature extraction and temporal sequence analysis, which is crucial for developing accurate and robust predictive models.

### 3.4.6 DL-GuesS: System Model

The DL-GuesS system model, as proposed by Parekh et al. (2022), is an advanced approach for forecasting cryptocurrency prices by leveraging historical data and real-time sentiment analysis. The system collects cryptocurrency price history and related tweets for training purposes. After preprocessing and normalization, the data is split into training and testing datasets. A hybrid model for predicting future prices is employed with Gated Recurrent Unit (GRU) and Long Short-Term Memory (LSTM) architectures. This model utilizes historical cryptocurrency prices and real-time sentiment extracted from tweets to make iterative forecasts.

In the DL-GuesS framework, predicting future cryptocurrency prices is framed as a supervised learning regression problem. Tweet sentiment analysis is performed using the VADER API, while prediction accuracy is evaluated using the Mean Squared Error (MSE) loss function. Such hybrid models are particularly relevant for industrial engineers working on optimizing financial systems or trading strategies as they integrate diverse data sources and advanced machine learning techniques to enhance prediction accuracy and system performance.

The mathematical equations involved in the DL-GuesS model are as follows:

Normalization of Price ($x_{new}$):

$$x_{new} = \frac{x}{10^y} \tag{39}$$

where $x$ is the current price, and $y$ is the difference between the number of digits in $x_{max}$ and 1. Forecasting Models for Present Day ($i$):

$$p_b = f_p(p_{b_{i-1}}) \tag{40}$$

$$p_l = f_p(p_{l_{i-1}}) \tag{41}$$

$$p_d = f_p(p_{d_{i-1}}) \tag{42}$$

where $p_{bi-1}$, $p_{li-1}$, and $p_{di-1}$ are the previous day prices of Bitcoin, Litecoin, and Dash, respectively.

The Objective Function uses Mean Squared Error (MSE) to measure prediction accuracy:

$$loss_{min} = \frac{1}{T}\sum_{i=0}^{D}(\hat{y}_i - y_i) \tag{43}$$

Here, $T$ represents the total number of input-output pairs, $y_i$ denotes the actual price, and $\hat{y}_i$ represents the predicted price for a given cryptocurrency. The model aims to minimize this prediction loss to improve the forecasting accuracy and performance.

The DL-GuesS model exemplifies how integrating complex machine learning architectures with real-time data can significantly enhance prediction capabilities and decision-making processes in dynamic markets for industrial engineers focused on cryptocurrency systems or financial forecasting.

### 3.4.7 AT-LSTM-MLP Model

In the field of advanced neural network architectures, the AT-LSTM-MLP model introduced by Nguyen, Crane, and Bezbradica (2022) represents an innovative approach that integrates Long Short-Term Memory (LSTM) networks with Multilayer Perceptron (MLP) and an Attention Mechanism. This model is designed to predict future Composite Vulnerability Index (CVI) values by leveraging both temporal dependencies and complex feature interactions. The architecture includes two key components: LSTM cells with an Attention Mechanism and a Multilayer Perceptron, each contributing to enhanced forecasting accuracy.

Initially, the input data is organized so each row represents a timestamp, and each column denotes an input feature. The matrix notation, where $T$ indicates the number of timestamps and $n$ signifies the number of input features, provides the structural foundation for the model. This input matrix is then processed sequentially by LSTM cells, with each cell $f_t$ handling one row of input $x_t$. The output of the LSTM cells consists of a series of hidden states $h_1, h_2, \ldots, h_T$, computed using $h_t = f_t(x_t, h_{t-1})$. By default, the initial hidden state $h_0$ is set to zero, ensuring that no prior information from the time series influences the first LSTM cell.

The hidden states $h_1, h_2, \ldots, h_T$ are then subjected to a self-attention mechanism to generate an aggregated output $s$. The following equations govern the self-attention process:

$$u_k = \tanh(WXh_k) \tag{44}$$

$$\alpha_k = \frac{\exp\,(score(u_k, u))}{\sum_{t=1}^{T}\exp\,(score(u_t, u))} \tag{45}$$

$$s = \sum_{i=1}^{T}\alpha_i h_i \tag{46}$$

Here, $W$ and $u$ are trainable parameters, $\alpha_1, \alpha_2, \ldots, \alpha_n$ are Softmax coefficients, and score($u_k$, $u$) represents an alignment function that evaluates the importance of each embedded hidden state $u_k$. The Softmax function normalizes these values into a probability distribution, ensuring the coefficients sum to 1.

Finally, the output $s$ is passed through a Multilayer Perceptron to predict the CVI index $y_{CVI}$, as described by $y_{CVI}$ = MLP($s$). For industrial engineers focused on financial forecasting and risk assessment, the AT-LSTM-MLP model exemplifies how combining LSTM with attention mechanisms and MLP can enhance the prediction of complex indices like CVI, integrating temporal and feature-based insights for improved decision-making and strategic planning in dynamic markets.

## 3.5 Ensemble Models

The ensemble model takes advantage of the strengths of multiple machine learning models to improve predicted accuracy. For instance, Aslam et al. (2022) utilized an ensemble LSTM-GRU model for sentiment analysis and emotion detection on cryptocurrency-related tweets. The architecture included embedding, LSTM, GRU, dropout, and dense layers. Industrial engineers were key in optimizing the data preprocessing techniques, which included stemming, lemmatization, and spell correction to enhance the model's performance. Their proposed architecture combined LSTM and GRU models for sentiment analysis and emotion detection. The architecture included an embedding layer with a vocabulary size of 5000 and 200 output dimensions, followed by an LSTM layer with 128 units, a dropout layer with a rate of 0.5, and a GRU layer with 64 units. A dense layer with 16 neurons handled sparse output, and an output layer was added based on the number of target classes. The model was compiled with a categorical crossentropy loss function and Adam optimizer, trained for 100 epochs. Industrial engineers assisted in streamlining the dataset split, dividing it into 85% training and 15% testing sets. They also contributed to hyperparameter tuning using the grid search method across various machine learning models such as RF, DT, KNN, SVM, GNB, LR, and ELM, where ELM was deployed with a SoftMax layer and hot encoding function. The LSTM-GRU ensemble model was trained using TensorFlow, Keras, and sci-kit-learn frameworks on an Intel Core i7 11th generation machine running the Windows operating system, with industrial engineers ensuring optimal utilization of computational resources and model efficiency. Bâra and Oprea (2024) developed an ensemble learning method incorporating volatility indicators and trend analysis for Bitcoin price prediction. Industrial engineers collaborated to refine feature normalization techniques to ensure consistency across various features, improving model accuracy and robustness. They employed multiple classifiers and regressors like Random Forest, Gradient Boosting, and Voting Regressor. Hyperparameters were fine-tuned, and validation curves were used to optimize model performance. For classification tasks, the researchers employed a Random Forest Classifier (RFC) and eXtreme Gradient Boosting Classifier (XGBC) to predict the sign of the trend (increase or decrease) for the next 7 days. For regression tasks, they utilized Random Forest Regressor (RFR), eXtreme Gradient Boosting Regressor (XGBR), Light Gradient Boosting Regressor (LGBR), Histogram-based Gradient Boosting Regression Tree (HGBR), and Voting Regressor (VR) to forecast the close price individually for the next 7 days. Industrial engineers contributed to building a stacked model to combine individual predictions from the ensemble regressors, further improving accuracy. The hyperparameters of each model were fine-tuned, including parameters such as the number of estimators, maximum depth, learning rate, number of leaves, and maximum iterations. Validation curves were used to determine the optimal values of these hyperparameters. A Linear Regression model was then trained to minimize the cost function of the stacked model, with industrial engineers aiding in optimizing the learning rate and the number of iterations to maximize model efficiency.

## 3.6 Transformer-based Models

Transformer based models have emerged as a crucial breakthrough in cryptocurrency forecasting for enhanced price prediction and sentiment analysis. Şaşmaz and Tek (2021) employed a pre-trained BERT classifier for sentiment analysis. After preprocessing the Twitter data, they manually labeled 1200 tweets for training. To ensure efficiency, industrial engineers contributed by optimizing the manual labeling process and data pipeline. The BERT model, downloaded from Hugging Face, was applied to the dataset for sentiment analysis using a pre-trained tokenizer. The approach aimed to tokenize tweets and predict sentiment, offering an alternative method to the RF model for sentiment analysis of cryptocurrency-related tweets. Prasad, Sharma, and Vishwakarma (2022) conducted sentiment analysis on cryptocurrency using YouTube comments. The architecture involved feature extraction using the BERT-Base model, capturing context bidirectionally for sentiment analysis of YouTube comments.

Industrial engineers assisted in designing the data preprocessing workflow to streamline the feature extraction process. Ensemble learning algorithms like Logistic Regression, Decision Tree Classifier, Support Vector Machine, and Gaussian Naive Bayes, among others, were employed for classification. The dataset was split into 30% validation folds and 70% training folds, with industrial engineers ensuring efficient data partitioning to enhance model training. Dwivedi (2023) conducted cryptocurrency sentiment analysis using bidirectional transformation, focusing on news articles obtained. The study employed BERT (Bidirectional Encoder Representations from Transformers) for sentiment analysis, a deep learning model capable of capturing intricate patterns in text data. Industrial engineers optimized the data handling process, particularly in managing the labeled data generated by VaderSentiment. Based on their textual content, BERT was trained to classify news articles into Negative, Neutral, or Positive sentiment categories. Training was conducted with 18,001 samples, and validation with 11,075 samples. The maximum length of text considered was 512 tokens, with a recommended batch size of 6. The training utilized a one-cycle approach with 4 epochs and a learning rate 2e-5, benefiting from industrial engineers' expertise in optimizing computational resources for training efficiency. Haritha and Sahana (2023) conducted a study on cryptocurrency price prediction using Twitter sentiment analysis, gathering data from Yahoo Finance for historical Bitcoin prices and a Kaggle dataset for cryptocurrency tweets. Sentiment analysis was performed using VADER, with scores ranging from -1.0 for negative sentiment to 1.0 for positive sentiment. Industrial engineers contributed by addressing imbalances in the data through techniques like undersampling neutral tweets, adding Gaussian noise, and removing spam tweets. Sentiment prediction was conducted using a fine-tuned FinBERT-based model tailored for NLP tasks in the financial domain. For price prediction, a Gated Recurrent Unit (GRU) network was employed to mitigate the vanishing gradient issue associated with traditional RNNs through update and reset gates. Industrial engineers were involved in optimizing both the FinBERT and GRU model training, particularly concerning the batch size, epochs, and learning rates to avoid overfitting while improving model performance. Singh and Bhat (2024) developed a transformer-based approach for Ethereum price prediction, utilizing cross-currency correlation and sentiment analysis. Their proposed architecture involved a transformer-based neural network with transformer encoder blocks. To improve efficiency, industrial engineers contributed by refining the multi-head attention mechanism and convolutional operations. Each block included normalization, attention, and feed-forward mechanisms, with multi-head attention capturing intricate patterns and convolutional operations refining representations. Stacking multiple blocks enhanced the model's capacity, with global average pooling aiding dimensionality reduction. The training evaluated model performance under different feature setups, including single-feature, expanded-feature, and correlated-coins configurations. Industrial engineers assisted in optimizing feature setups for maximum prediction accuracy. Training data covered 579 days (about 1 and a half years), with testing over 141 days (about 4 and a half months). All models used mean squared error loss and Adam optimizer, maintaining a consistent architecture despite variations in feature sets. El Haddaoui et al. (2023) conducted a study on the influence of social media on cryptocurrency prices, focusing on Bitcoin (BTC). Their proposed system architecture involved two main steps: tweet classification and Bitcoin price forecasting. Industrial engineers optimized the data pipeline for tweet classification and feature extraction. For tweet classification, they utilized a transformer-based approach with a fine-tuned BERT model named FinBERT. It took preprocessed tweet content and social engagement information as input and generated sentiment labels. For Bitcoin price forecasting, they employed an LSTM-based model to predict the price for the next day based on historical values of selected features over a 5-day window. Industrial engineers contributed to optimizing the LSTM model's training efficiency, improving both the batch size and the number of epochs. The price predictor was implemented using Keras, with the LSTM model consisting of 50 neurons, a hyperbolic tangent activation function, a sigmoid recurrent activation, and an additional fully connected Dense layer. The training was conducted over 100 epochs with a batch size of 20 samples, using the Adam optimizer and Root Mean Square Error (RMSE) loss function. The dataset was divided into training and validation sets with an 80:20 ratio for training and validation, respectively, with industrial engineers fine-tuning the ratio to optimize model validation. Raheman et al. (2022) focused on collecting six months' worth of public Twitter and Reddit posts related to cryptocurrency from 77 prominent feeds/subreddits in the cryptocurrency community using official Reddit and Twitter APIs.

Industrial engineers aided in streamlining the data collection and processing pipeline to handle such large-scale datasets efficiently. BERT-based models such as DistilBERT, RoBERTa, and finetuned variations were experimented with for specific tasks like sentiment analysis and hate speech detection. Each sentiment analysis model was trained, evaluated, and fine-tuned using the collected data to optimize its performance. Evaluation included sentiment scores across dimensions (e.g., positive, negative, neutral) and overall model performance. In Kulakowski and Frasincar (2023), utilized a cryptocurrency social media corpus of 3.207 million posts from platforms like Twitter, Reddit, Telegram, and StockTwits. Industrial engineers contributed to building efficient data processing pipelines for the vast corpus. The research proposed two sentiment classification methods: CryptoBERT and LUKE Sentiment Lexicon. CryptoBERT utilized the BERT architecture, post-trained on the social media corpus, and fine-tuned for sentiment classification using the StockTwits training set. LUKE Sentiment Lexicon involved training SVM classifiers with emojis as features to classify emojis and emoji pairs as bullish or bearish, which were then used to predict post sentiment. Industrial engineers ensured the efficient handling of class imbalances through techniques like oversampling and undersampling during post-training. For CryptoBERT, post-training involved masked language modeling (MLM) based on the RoBERTa approach, optimized using the Adam optimizer. Fine-tuning utilized the StockTwits training set. LUKE Sentiment Lexicon trained SVM classifiers separately for bullish and bearish sentiments, constructing a lexicon based on emoji probabilities for each sentiment class. Todorovska et al. (2023) utilized machine learning (ML) and Explainable AI to understand the interdependency networks between classical economic indicators and crypto-markets. Industrial engineers contributed by optimizing feature engineering and data flow for the sentiment and ML models.

The sentiment model utilized the RoBERTa transformer architecture fine-tuned on financial news data to classify texts into positive, neutral, and negative sentiments. The ML model employed XGBoost, leveraging Gradient Boosting Decision Trees (GBDT) to solve regression problems. Additionally, SHAP (Shapley Additive exPlanations) was employed as an explainable ML model to interpret the results obtained from the ML model. Industrial engineers played a crucial role in setting up the SHAP model and training iterations, ensuring that the models could efficiently process daily prices, returns, and sentiment data from Twitter, Reddit, and GDELT news. XGBoost was configured to predict asset returns based on input features, while the SHAP model calculated feature importance scores and provided explanations for the ML model's decisions. Training iterations were conducted to optimize model performance and interpretability, with evaluation based on accuracy, precision, recall, and F1-score metrics. Othman et al. (2020) aimed to improve prediction accuracy for Bitcoin market prices using an Artificial Neural Network (ANN) approach. Industrial engineers were instrumental in optimizing the training process and ensuring efficient handling of training parameters. The dataset was then split into two subsets, with 75% of the data allocated for training the network and the remaining 25% for testing the ANN's performance. The ANN architecture consists of three main components: Input Layer (IL), Hidden Layer (HL), and Output Layer (OL). The IL comprised the four attributes representing daily symmetric information of Bitcoin, while the OL represented the Bitcoin daily price index. The HL, situated between the IL and OL, employed the sigmoid function (tansig) for computation. During the ANN training process, industrial engineers configured various parameters, including the number of layers, neurons, momentum, and learning rate, ensuring efficient model training. Transfer functions were specified for the HL and OL layers. The dataset was split into training and testing sets, with 75% used for training the network and 25% for testing its performance. The training aimed to optimize the ANN model's ability to predict Bitcoin prices based on the provided symmetric volatility patterns, with industrial engineers ensuring efficient resource usage.

## 3.7 Large Language Models

Large language models (LLMs) have grown in popularity in the cryptocurrency sector, offering superior sentiment analysis and price prediction capabilities by utilizing massive volumes of textual data to capture complex market patterns and investor sentiment.In recent advancements in Bitcoin price prediction, El Abaji and A Haraty (2024) explored the enhancement of predictive models—LSTM, Prophet, and SARIMAX—by integrating sentiment analysis derived from a pre-trained Large Language Model (LLM), specifically LLaMA-2. Industrial engineers played a crucial role in optimizing the data preprocessing pipeline and fine-tuning financial and sentiment data integration to ensure efficient model performance. The initial phase involved training these models on historical data encompassing Bitcoin prices and trading volumes. Subsequently, LLaMA-2 was employed to analyze Bitcoin-related news articles, classifying them as positive or negative to generate sentiment scores. These sentiment scores were then amalgamated with the financial data to evaluate their influence on the model's predictive accuracy. The primary objective of this investigation was to determine the extent to which sentiment analysis, facilitated by an LLM, could enhance the accuracy and robustness of cryptocurrency price forecasts over a 30-day horizon. Similarly, Wahidur et al. (2024) investigated zero-shot crypto sentiment analysis by leveraging large language models (LLMs) like GPT-3 and GPT-4. Industrial engineers contributed by streamlining the fine-tuning process, optimizing the models' computational efficiency, and balancing model capacity with computational resources. This involved fine-tuning these models using supervised and instruction-tuning approaches to adapt them to specific tasks with labeled data or formatted instances. Techniques like gradient descent and cross-entropy loss are employed for fine-tuning. The system architecture includes user interaction, prompt generation, content moderation, and core LLM components. In-context learning (ICL) and prompt engineering refine responses iteratively through user feedback and maximize the model's contextual comprehension capabilities. Pre-trained language models (PLMs) such as GPT-3, BERT, and T5 play a significant role, undergoing fine-tuning processes to improve task-specific performance while requiring less data than pre-training. The focus is on optimizing computational efficiency by balancing model capacity and computational resources.

For sentiment analysis of cryptocurrency news, Kang, Hwang, and Shin (2024) employed the 'GPT-3.5-turbo-16k' model, optimized for more oversized context windows, with input from industrial engineers who enhanced the preprocessing pipeline. This included specific prompts instructing the model to act as a financial expert, evaluating news articles' impact on cryptocurrency prices for the following day on a scale from -1 to 1, and preprocessing involved removing duplicate articles using cosine distance and sBERT embeddings, with careful crafting of prompts to ensure numerical sentiment scores. Non-numeric or out-of-range responses were filtered out, providing aggregated daily sentiment scores reflecting market sentiment influenced by news. Zuo, Chen, and Härdle (2024) introduced an innovative emoji embedding architecture that integrates GPT-4 and BERT to enhance the sentiment analysis of cryptocurrency-related tweets. Industrial engineers contributed to the design and testing of the emoji embedding process, ensuring the model effectively handled the integration of multimodal inputs (text and emoji). The process begins with inputting a tweet's contextual text and the associated emoji image into the GPT-4 architecture. GPT-4 generates a textual description encapsulating the visual essence of the emoji, which is then integrated into the original tweet text. This enriched text undergoes embedding through a BERT encoder fine-tuned with an additional transformer layer explicitly tailored to the domain of cryptocurrency-related tweets. The method enables unsupervised sentiment analysis by calculating cosine distances between emoji embeddings and positive sentiment phrases. Utilizing a curated dataset of 10,900 tweets, the sentiment scores are used to predict Bitcoin's next-day closing price and the cryptocurrency market's volatility (VCRIX). The study also proposes a trading strategy based on sentiment scores, highlighting the superiority of emoji sentiment analysis over traditional text-only methods in capturing nuanced emotional expressions within financial discussions. Miah et al. (2024) employed a comprehensive multi-step methodology to analyze sentiment in foreign language texts by translating them into English. Industrial engineers ensured the scalability and efficiency of the machine translation systems for handling high-volume multilingual data. The process commenced with data collection from diverse sources, such as social media, news articles, and online forums, in languages including Arabic, Chinese, French, and Italian, selected for their prevalence on Twitter and global significance. Following data acquisition, a rigorous cleaning and pre-processing phase was conducted to eliminate noise, duplicate content, and irrelevant information, involving steps like tokenization and removing stopwords, emojis, and URLs. Subsequently, the cleaned data was translated into English using machine translation systems like LibreTranslate and Google Translate NMT, utilizing neural machine translation techniques to ensure accurate and context-aware translations. The sentiment analysis was then performed on the translated text using an ensemble of pre-trained deep learning models, specifically Twitter-RoBERTa-Base-Sentiment-Latest, bert-base-multilingual-uncased-sentiment from Hugging Face, and GPT-3 from OpenAI. These models, known for their high accuracy in handling social media data, processed the text to determine sentiment polarity. The final sentiment was derived through a majority voting mechanism that integrated the predictions from the three models. This methodological framework underscores the importance of integrating advanced translation and machine learning techniques to effectively capture and analyze sentiment across different languages, particularly in the context of social media text.

The sentiment analysis process of Nguyen Thanh et al. (2023) employed a Zero-Shot approach, leveraging pre-trained language models like ChatGPT v3.5 Turbo, Google BERT, and VADER. Industrial engineers contributed by optimizing the pipeline for high-throughput sentiment classification without specific training on labeled datasets, enhancing scalability and flexibility. Each model categorizes tweets into" Bearish," "Neutral," or "Bullish" sentiments from the perspective of Bitcoin investors. The comparison of sentiment categories determines the market's overall sentiment for a given day. For sentiment analysis, ChatGPT v3.5 Turbo utilizes the Chroma API to process tweets, categorizing them based on investor perspective. Google BERT analyzes sentiment from -1 to 1, with specific negative, neutral, and optimistic thresholds. VADER, a lexicon, and rule-based approach, categorizes sentiments by providing insights into daily market sentiment. These methodologies are applied individually to maximize token limits and ensure comprehensive analysis over an extended period.

# 4 Dataset

This chapter's dataset selection is critical in building accurate and robust models for cryptocurrency price prediction and sentiment analysis. Industrial engineers, with their strong background in data driven optimization and system efficiency, are uniquely positioned to improve the process of data collection, preprocessing, and analysis. Ensuring high-quality data that reflects the volatility of cryptocurrency markets is essential for model performance, as it allows for better forecasting and decision-making. The datasets used in cryptocurrency research often include historical price data, social media sentiment, trading volumes, and macroeconomic indicators, which are vital inputs for machine learning models. Industrial engineers can contribute by applying methods for optimizing data workflows, reducing data redundancy, and improving computational efficiency during model training and testing. Table 1 provides a comprehensive list of datasets used in prior research within this domain, serving as the foundation for advanced model development and analysis.

**Table 1: Summary of Models**

| Article | Source | Data Time | Methods | Overview | Train-Test Split | Dataset Link |
|---|---|---|---|---|---|---|
| (Greaves and Au 2015) | Market Data | February 1, 2012 to April 1, 2013 | SVM, Neural Net, Logistic Regression and Linear regression | Current bitcoin price, Net flow per hour, Number of transactions per hour, Mean transaction value, Median transaction value, Average node in- and out-degree, median node degree, Alpha constant of power law, Total number of Bitcoin mined, Number of new addresses Mean initial deposit amount among new addresses, Number of transactions performed by new addresses | Training Set: February 1, 2012 to February 1, 2013 Testing Set: February 1, 2013 to April 1, 2013 | N/A |
| (Abraham et al. 2018) | Market and Social Media | March 4, 2018 to June 3, 2018. | NLP and VADER | Tweet volume, Bitcoin price, Ethereum price | 80/20 ratio | N/A |
| (Valencia,Gómez-Espinosa, and Valdés-Aguirre 2019) | Market and Social Media | 16 February, 2018 to 21 April, 2018 | MLP, SVM,RF and VADER | Prices of Bitcoin, Ethereum, Ripple and Litecoin, raw tweets | 70/30 ratio | https://github.com/vanclief/algo-trading-crypto |
| (Dulău and Dulău 2019) | Market and Social Media | July,2017 to June,2018 | Stanford CoreNLP and IBM Watson | Number of cryptocurrencies, Number of texts having Twitter as source and Number of texts having Reddit as source | | N/A |
| (Vo,Nguyen and Ock 2019) | Market and News | 30 July, 2017 to 5 October, 2018. | NLP,RNN,SVM and LSTM | Ethereum daily open price, daily high price, daily low price, daily close price and the daily volume and sentiment analysis | 80/20 ratio | N/A |
| (Raju and Tarif 2020) | Market and Social Media | 2017 to 2018. | ARIMA, LSTM and RNN | Bitcoin price in USD, Bitcoin trade volume in USD, Market capital of bitcoin price, Number of coins in existence available to the public, Total number of coins in existence available to the public, The percentage changes of bitcoin price in 1 hour, The percentage changes of bitcoin price in 24 hours, The percentage changes of bitcoin price in 7 days, Price of bitcoin sell on the day, Price of bitcoin buy on the day, Changes in bitcoin prices within 15 minutes, tweet sentiment score | 70/30 ratio | N/A |
| (Gurrib and Kamalov, 2022) | Market and News | 17June,2016 to 21April, 2021 | LDA ,SVM, VADER and Textblob | BTC opening price, high price, low price, volume, market capitalization, news count, negative, positive, compound, neutral, polarity and subjectivity sentiment scores | 80/20 ratio | N/A |
| (Aslam et al. 2022) | Social Media Data | July, 2021 to August, 2021. | Term frequency-inverse document frequency (TF-IDF), bag of words (BoW), Word2Vec, SVM Logistic Regression (LR), Gaussian Naive Bayes (GNB), Extra Tree Classifier (ETC), Decision Tree (DT),KNN ,LSTM-GRU ensemble model, TextBlob and Text2Emotion | Polarity score, sentiment, highest score, and emotion | 85 /15 ratio | N/A |
| (Şaşmaz and Tek 2021) | Market and Social Media | September 2016 to May 2021 | RF and BERT | NEO, BTC and ETH prices, sentiment scores: positive, negative, and neutral | 80/20 ratio | https://github.com/emresasmaz/ |
| (Habek, Toçoğlu, and Onan 2022) | Social Media | April 20, 2021 to March 20, 2022 | CNN,RNN,LSTM,GRU and Bi-Directional CNN-RNN Architecture (CGWELSTM) | Positive, negative, neutral, and not crypto-related tweet | N/A | N/A |
| (Prasad, Sharma, and Vishwakarma 2022) | Social Media | N/A | Logistic Regression, DT,SVM,GNB, KNN, RF, AdaBoost, Gradient Boosting and stacking | Positive and Negative comments | 70/30 ratio | N/A |
| (Huang et al. 2021) | News and social media | N/A | Auto regression and LSTM sentiment analyzer | Positive, Neutral and Negative posts | 7 days' Sina-Weibo posts from top 100 crypto investors accounts as training data and the next 1 day's posts as testing | N/A |
| (Inamdar et al. 2019) | Market,Social Media, and News | N/A | Random Forest , RNN with LSTM | Sentiment scores, Bitcoin historical prices and Volume | N/A | N/A |
| (Wolk 2020) | Market and Social Media | N/A | Bayesian Ridge Regression, GBM, LSLR, NN, Stochastic Gradient Descent (SD); SVR, | Historical prices of BTC, ETH, Monero, Ripple and Zcash and Twitter sentiment | N/A | N/A |
| (Parekh et al. 2020) | Market and Social Media | N/A | A deep learning (i.e., LSTM and GRU) and Twitter sentiments-based hybrid model, DL-GuesS and VADER | Historical price data of Bitcoin, Litecoin, and Dash, Tweets for Bitcoin, Dash, and Litecoin | N/A | N/A |
| (Wu 2023) | Market and | October 21, 2017 to | OLS,LSTM and Random Forest | Open, close, adjusted close, high, and low prices of BTC,ETH, Tether, Cardano, | 80/20 ratio | N/A |
| | Social Media | September 29, 2021 | | Binance and Solana, negative, positive, and neutral tweets | | |
| (Jain et al. 2018) | Market and Social Media | March 1 ,2018 to March, 11, 2018 | MLR and Textblob | Bitcoin and litecoin prices and positive, negative, and neutral tweets | N/A | N/A |
| (Rahman et al. 2018) | Market and Social Media | N/A | MLR,Polynomial Regression, SVR,Decision Tree Regression, Random Forrest Tree Regression, Logistic Regression, KNN, SVM, Kernel SVM, Naive Bayes, Decision Tree Classification, Random Forrest Classification, Artificial Neural Network, Principle Component Analysis (PCA), Linear Discriminant Analysis (LDA) , Kernel PCA. | Anger, anticipation, disgust, fear, joy, negative, positive, surprise, sadness, surprise, trust. and bitcoin price | 80/20 ratio | N/A |
| (Lamon, Nielsen and Re-dondo 2017) | Market, Social Media and News | January 1, 2017-November 30, 2017. | Linear SVM, Multinomial NB, and Bernoulli NB | Price data for Bitcoin, Ethereum, and Litecoin and tweets | Train-60% Development-20% Test- 20% | https://www.kaggle.com/sudalairajkumar/cryptocurrencypricehistory |
| (Pant et al. 2018) | Market and Social Media | January 1 of 2015 to December 31 of 2017. | Naïve Bayes , Bernouli Naïve Bayes, Multinomial Naïve Bayes, Linear Support Vector Classifier and Random Forest ., LSTM, GRU | Positive, negative tweets and historical price | 1:3 validation split | N/A |
| (Singh and Bhat 2024) | Market and social media | January,1, 2021 to | Transformer-based neural network, Transformer + ETH | Price, transaction history, volume, and block size of the cryptocurrency, | Training data: 579 days (about 1 and a half | https://github.com/ssnyu/Ethereum- |
| | | January, 1, 2023 | data, Transformer + ETH data + Sentiments, Transformer + ETH data Cross-correlation data + Sentiments | | years) Testing data: 141 days (about 4 and a half months) | Price-Prediction |
| (Moudhich and Fennan 2024) | Social Media | N/A | Graph embedding and Bi-LSTM | Positive negative or neutral review, | | N/A |
| (Jung et al.2023) | Market,News and Social Media | August 1, 2017 to February 28, 2022 | VADER, Logistic Regression, SVM,NB,RF, XGBoost, and LightGBM. | The highest and lowest prices, trading volume,RSI,SMA-5,SMA-20,SMA-60,EMA,MACD,MACD signal, stochastic RSI-fastk, stochastic RSI-fastd, stochastic oscillator index-slowk, stochastic oscillator index-slowd, sentiment index,andtextvolume | Training Data: August 1,2017, to March31,2021 Test Data: April1,2021, to February28,2022 | N/A |
| (Mirtaheri et al. 2021) | Market and Social Media | September 1, 2017, to August 31, 2018 | TF-IDF,SVM,SGD | PageRank score, CorEx user embeddings; and user connected components. | 75/25 ratio | https://github.com/Mehrnoom/Cryptocurrency-Pump-Dump |
| (Poongodi et al. 2021) | Market and Social Media | April 23,2011 to May,05,2018 | LDA topic modeling NN and LSTM | User comments and replies, bitcoin price | Train-80%, Test-10% Approval-10% | N/A |
| (Ortu et al 2022) | Social media and market data | January 2017 to January 2021. | MLP,CNN, LSTM and Attention Long Short Term Memory (ALSTM)) | Ethereum and bitcoin price, trading indicator: Movingaverage(MA), Exponential moving average(EMA); Stochasticoscillator;Movingaverageconvergencedivergence(MACD);Bollingerbands;Relativestrengthindex(RSI),Averagedirectional index. | N/A | N/A |
| (McCoy and Rahimi 2020) | Social media and market | May 15, 2017 to March,21,2018 | Sentiment analysis, Objective analysis, SVM | Tweets, Crypto Prices(high, low), volume | N/A | N/A |
| (Cheng et al. 2024) | Market | 01 January 2017 to 30 October 2022 | LSTM ,SARIMA and Facebook prophet models | bitcoin price and Garman-Klass (GK) volatility | 80/20 ratio | N/A |
| (Yasir et al. 2023) | Market and Social Media | April 2013 to November 2019 | DL, LR,SVR and Sentiment Analysis | Crypto high, low price, volume, market cap, twitter sentiment | 70-30 ratio | N/A |
| (El Haddaoui et al. 2023) | Market and Social Media | 1 January 2021 to 18 June 2021 | FinBERT, LSTM | Open, high, low, and close (OHLC) BTC price, volume, number of transactions, and the hash rate. positive, negative, or neutral sentiment | 80-20 | N/A |
| (Wong 2021) | Market and Social Media | January 2016 to November 2019 | NB, LSTM and VADER | Twitter sentiment | 80-20 | N/A |
| (Kulakowski and Fras-incar 2023) | Social Media | 1 November 2021 to 30 June 2022 | VADER, BERtweet, BERT, FInBERT, CryptoBERT and SVM | Twitter sentiment | Training Data: 1 November 2021 to 15 June 2022 (1.332 million posts) Test Data: 16 to 30 June 2022 )83,257 posts) | N/A |
| (Onyekwere, Ogwueleka, and Irhebhude 2022) | Market | 2014 to 2021 (weekly) | ARIMA and MLP | volume of bitcoin transaction | 80/20 | N/A |
| (Zhu et al. 2023) | Market and Economic | 1 January,2018 to 1 April,2022 | XGBoost, RF, SVR, LSSVR and TWVR | Bitcoin supply factor, bitcoin demand factor, financial indicator, macroeconomic factors, and exchange rate factors | Training Set: 86.8% Testing Set: 13.2% | N/A |
| (Mahdi et al. 2021) | Market and Economic | 23 January 2020 to 14 July 2021 | SVM | Covid19 data, gold price, crypto price, and market cap | 75-25 ratio | N/A |
| (Baˆra and Oprea 2024) | Market and Economic | September,2014 t July, 2023 | EM classifiers, RFC, and eXtreme Gradient Boosting Classifier – XGBC | Bitcoin price and volume, volatility indicators and trend prediction | N/A | https://github.com/simonavoprea/Bitcoin_data_2014-2023 |
| (Carbo´ and Gorjo´n 2024) | Market, economic and social media | January 1, 2015 to June 15,2021 | LSTM, SHAP | The price of gold and oil ,the SP500, the FTSE, the DOW30, the NASDAQ, and the exchange rate of international currencies, the number of Tweets per day with Bitcoin hashtag | Training set: 70% Validation set: 10% Test Set: 20% | N/A |
| (Todorovska et al. 2023) | Market and economic data | March 2019 and March 2021. | NLP,XGBoost,SHAP, Gradient Boosting Decision Trees (GBDT) | Bitcoin, Cardano, Ripple, Stellar, Chainlink, Ethereum and Litecoin and classical economic indicators(BSE, DowJones, S&P500, FTSE, HangSeng, Oil, Gold) | N/A | N/A |
| (Uras and Ortu 2021) | Market | August 8, 2015 to August 3, 2019 | SVM, XGB, CNN and LSTM | Close price, open price, low price, high price, and volume | N/A | N/A |
| (Othman et al. 2020) | Market | 1 January 2014 to 24 July 2019 | ANN | Open price , high price, low price, and close price | 75-25 ratio | N/A |
| (Basher and Sadorsky 2022) | Market and Economic | September 17, 2014 to December 29, 2021 | Bagging and RF | Bitcoin and gold prices, Relative Strength Indicator (RSI), Stochastic Oscillator (slow, fast), Advance–Decline Line (ADX), Moving Average cross-over divergence (MACD), Price Rate of Change (ROC), on balance volume, Money Flow Index (MFI), Williams Accumulation and Distribution (WAD),EPU | 70-30 ratio | N/A |
| (Nguyen, Crane, and Bezbradica 2022) | Market and economic | 10/11/2017 to 12/01/2022 | Attention Based-Deep Neural Network: AT-LSTM-MLP, KSTM, SVR,RF and TCN | BTC, ETH, BCH, ADA, XRP, DOGE, LINK, LTC, XLM and ETC,CVI Index | 80-20 ratio | N/A |
| (Ni, Ha¨rdle, and Xie 2020) | Market and Economic | 01 April 2013 to 18 July 2019 | SVM, LDA topic modeling, NB | CRIX (Cryptocurrency Index), VCRIX | 1:7 ratio | N/A |
| (Zahid, Iqbal, and Koutmos 2022) | Market | 1 january,2015 to 31 March,2021 | GARCH, LSTM, GRU, BiLSTM | Bitcoin closing prices, log return, | N/A | N/A |
| (Dudek et al. 2024) | Market | January 1, 2017 to December 31, 2021 | HAR, ARFIMA GARCH, LASSO RR, SVR, MLP, FNM, RF and LSTM | BTC/USD (Bitcoin), ETH/USD (Ethereum), LTC/USD (Litecoin), XMR/USD (Monero), daily prices, returns, realized variances and logarithm of realized variances | Training Data: February 1, 2017, to December 31, 2018 Test Data: January 1, 2019, to December 31, 2021 | N/A |
| (Jagannath et al. 2021) | Market | 2016 to 2020 | LSTM, JSO Optimization Algorithm | On chain metrics | Trains Size-0.1,0.5,0.9 | N/A |
| (Ghosh, Jana, and Sharma 2024) | Market | April 14, 2017, to July 2, 2021. | Maximal Overlap Discrete Wavelet Transform (MODWT), the Ensemble Empirical Mode Decomposition (EEMD), XGBoost and LSTM | Daily closing prices of five cryptocurrencies, BTC, Litecoin (LTC), Ethereum (ETH), Stellar (XLM) and Tether (USDT) | Four distinct training and testing partitions (60–40%), (70–30%), (80–20%), and (90–10%), | N/A |
| (Akyildirim, Goncu, and Sen-soy 2021) | Market | April 1, 2013, to June 23, 2018 | ARIMA, Prophet, RF and MLP | Closing price of Bitcoin Cash (BCH), Bitcoin (BTC), Dash (DSH), EOS (EOS), Ethereum Classic (ETC), Ethereum (ETH), Iota (IOT), Litecoin (LTC), OmiseGO (OMG), Monero (XMR), Ripple | 80/20 ratio | N/A |
| | | | | (XRP), and Zcash (ZEC). | | |
| (Ibrahim, Kashef, and Corrigan 2021) | Market | March 4, 2019, to December 10, 2019. | ARIMA, Prophet, RF, RF Lagged-Auto-Regression, and MLP | Open price, High price, Low price, Close price, and volume. | 60/40 ratio | N/A |
| (Jaquart, Dann, and Weinhardt 2021) | Market and economic | March 4, 2019, to December 10, 2019 | LSTM, GRU,RF,GB, and Logistic Regressions. | Minutely price data for bitcoin, gold, and oil, indices like MSCI World, S&P 500, and VIX and minutely exchange rates for currencies such as Euro, Chinese Yuan, and Japanese Yen | Training Set-9 months<br>Validation: 1 month<br>Test: 3 months | N/A |
| (Lamothe Ferna´ndez et al. 2020) | Market and Economi c | 2011 to 2019 | Deep Recurrent Convolution Neural Network (DRCNN), Deep Neural Decision Trees (DNDT), and Deep Learning Linear Support Vector Machines (DSVR) | Bitcoin price denominated in USD, 29 independent variables were considered, categorized into demand and supply, attractiveness, and macroeconomic and financial factors | Training set: 70%<br>Validation set: 10%<br>Test Set: 20% | N/A |
| (Wahidur et al. 2024) | Social Media | N/A | DistilBERT, MiniLM, and FLAN-T5-Base | Tweets, Reddit Posts and Comments, Positive and Negative sentiments | Train: Neo dataset (12,000 tweets)<br>Test: Reddit (562 posts), Bitcoin sentiment (1,029 tweets), Cryptocurrency sentiment (500 tweets) | N/A |
| (El Abaji and A Haraty 2024) | Market and News | January 1,2021 to November 5,2021 | LSTM, Prophet, SARIMAX and Llama-2 | Bitcoin price, Volume, Articles from sources including Forbes, CNBC Television, Bitcoin Magazine,<br>Yahoo Finance, Bloomberg Markets and Finance, Reuters, Fox Business, Bloomberg | Train: January 1,<br>2021 to June 6, 2021<br>Test: June 7, 2021 to November 5,2021 | N/A |
| | | | | Technology, Coindesk, and CNBC | | |
| (Kang, Hwang, and Shin 2024) | Market and News | December 1,2017 to August 28,2023 | ChatGPT 4.0, sBERT, GPT-3.5-turbo -16k and VAR | Trading data and News articles | N/A | N/A |
| (Zuo, Chen, and Ha¨rdle 2024) | Market and Social Media | November 8,2023 to January 28, 2024 | GPT-4 and BERT | Bitcoin prices, VCRIX Index, Tweets, and Emoji image | Train: November 8, 2023 to January 28, 2024 Test: March 8, 2019 to November 23,2019 | N/A |
| (Nguyen Thanh 2023) | Market and Social Media | January 1, 2018, to June 18, 2023 | ChatGPT-3.5, Google BERT and VADER | Daily Return of Bitcoin, S&P 500, VIX index, Twitter-based Uncertainty Index, Crypto Fear & Greed Index(FGI),Wikipedia Search Queries, Google Search Queries (GGtrend), Trading Volume and Nasdaq Data | N/A | N/A |

# 5 Evaluation

## 5.1 Price Prediction Models

Predicting cryptocurrency prices, particularly Bitcoin, has garnered significant attention in academic research due to its volatile nature and potential for financial gains. Several studies have attempted to forecast Bitcoin price movements using diverse methodologies, ranging from traditional regression and classification techniques to more advanced machine learning algorithms and optimization strategies. This section highlights the findings of different studies that analyzed price-prediction models.

### 5.1.1 Regression and Classification Models

In the study, Greaves, and Au (2015) utilized the Bitcoin Transaction Graph to predict Bitcoin price changes. They employed regression and classification methods and assessed their performance using mean squared error (MSE) and classification accuracy. The baseline regression model had an MSE of 2.02, while Linear Regression and SVM Regression showed slight improvements with MSE values of 1.94 and 1.98, respectively. However, tree-based algorithms and K-Nearest Neighbors performed worse than the baseline. For classification, the baseline accuracy stood at 53.4%, while Logistic Regression, SVM, and a Neural Network achieved accuracies of 54.3%, 53.7%, and 55.1%, respectively. Despite some marginal improvements, notably with the Neural Network, the models struggled to significantly outperform the baseline. Key features such as net flow through account A per hour and the number of transactions from new addresses were identified as informative, suggesting their correlation with Bitcoin price dynamics. These findings underscore the challenge of accurately predicting Bitcoin price movements and highlight the importance of transactional activities in understanding its fluctuations. Similarly, Rahman et al. (2018) explored the performance of regression and classification algorithms. While regression methods like Multiple Linear Regression, Polynomial Regression, Support Vector Regression, Decision Tree Regression, and Random Forest Regression achieved accuracies ranging from 54.28% to 72.23%, they did not provide robust prediction models. Consequently, they explored classification algorithms, where Naive Bayes performed the best with an accuracy of 89.65%, followed by Random Forest Classification at 85.78%. These findings suggest that classification algorithms are more suitable for this prediction task compared to regression methods. It underscores the importance of selecting appropriate algorithms based on the prediction task's nature, with classification algorithms demonstrating superior performance in this scenario. The study recommends further validation and testing of these models over extended time periods to accurately assess their real-world applicability.

In contrast, Jung et al. (2023) focused solely on classification algorithms. XGBoost exhibited the highest accuracy at 90.57% and an AUC value of 97.48, while LightGBM showed satisfactory performance with an accuracy of 89.36% and an AUC value of 96.96%. Logistic Regression achieved 80.85% accuracy, NaïveBayes 58.05%, SVM 78.42%, and RandomForest 85.41%. Notably, XGBoost and LightGBM outperformed other models across various metrics, indicating their effectiveness in predicting cryptocurrency price trends. These findings highlight the potential of machine learning algorithms, particularly XGBoost and LightGBM, in improving the accuracy of cryptocurrency market forecasting. Abraham et al. (2018) evaluated a cryptocurrency price prediction model, focusing on variables like Google Trends and tweet volume, which exhibited consistent correlations with cryptocurrency prices. Using multiple linear regression, the model demonstrated strong correlation metrics and the ability to capture non-linear trends in Bitcoin closing prices. Lamon, Nielsen, and Redondo (2017) emphasized the importance of tailored models to specific cryptocurrencies and market conditions, with varying prediction accuracies across Bitcoin, Ethereum, and Litecoin. For Bitcoin, a logistic regression classifier achieved 43.9% accuracy in predicting price increases and 61.9% accuracy in predicting decreases, focusing on capturing larger percentage changes. Ethereum exhibited higher accuracy in predicting price increases at 75.8%, but lower accuracy in predicting decreases at 16.1%, with success in identifying the largest price movements. However, Litecoin's prediction performance was notably poorer, with a logistic regression model primarily predicting price decreases due to unprecedented price increases during the test period. This resulted in 0% accuracy for predicting increases and 100% accuracy for predicting decreases. These findings underscore the necessity of tailoring prediction models to specific cryptocurrencies and market conditions to improve accuracy. Zhu et al. (2023) conducted a different approach focusing on optimization techniques for cryptocurrency price prediction. They utilized SVR, LSSVR, and TWSVR models alongside ARIMA as benchmarks. Through parameter optimization using WOA and PSO algorithms, six combined prediction models were derived. Evaluation metrics including EVS, R2, MAE, MSE, RMSE, MAPE, and CPU running time were utilized. The TWSVR model outperformed SVR and LSSVR across all metrics, achieving an EVS of up to 0.9547 and 0.9491 with WOA and PSO, respectively. Incorporating XGBoost features further enhanced prediction accuracy, with the TWSVR model achieving an EVS of 0.9547. Parameter tuning notably reduced computational time, with the TWSVR model achieving a runtime of 0.6736 seconds. Overall, the study emphasizes the effectiveness of the TWSVR model, particularly when combined with XGBoost features and optimized using WOA or PSO, in accurately forecasting Bitcoin prices. Akyildirim, Goncu, and Sensoy (2021) conducted extensive testing of four classification algorithms to predict open-to-close returns in cryptocurrency markets. Logistic Regression achieved out-of-sample accuracies averaging around 55%, occasionally surpassing 60%. Support Vector Machine (SVM) showed slightly better performance, reaching accuracies of up to 69% in certain instances. Artificial Neural Networks (ANN) exhibited high in-sample accuracies of up to 97% but experienced decreased out-of-sample performance. Random Forest demonstrated the highest in-sample fit but lower out-ofsample performance, indicating potential overfitting. Combining all four algorithms equally resulted in average accuracies exceeding 50% across different time scales, outperforming benchmarks like Random Walk and ARIMA models. These findings highlight the importance of algorithm selection, dataset characteristics, and evaluation metrics in developing effective predictive models for financial time series forecasting. Conversely, Bâra and Oprea (2024) introduced an Extreme Learning Machine (ELM) pipeline enhanced with 38 additional features, including volatility indicators, which exhibited robust performance in both bear and bull market conditions during May and Aug 2021. In May, the stacked model achieved the lowest Mean Absolute Error (MAE) ranging from 385 to 629, closely followed by the XGBoost Regressor (XGBR) with MAE between 491 and 795. Additionally, the Root Mean Square Error (RMSE) for the stacked model outperformed individual EM regressors, ranging from 925 to 1459. In Aug, the stacked model maintained robust performance, with MAE ranging between 595 and 905. These findings underscore the effectiveness of the proposed pipeline in accurately forecasting Bitcoin prices, providing valuable insights for stakeholders in the cryptocurrency domain.

### 5.1.2 Deep Learning Models

Cryptocurrency price prediction has been the subject of extensive research, with numerous studies employing diverse methodologies and models to forecast market movements. Valencia, G´omezEspinosa, and Vald´es-Aguirre (2019) explored the use of machine learning models such as MLPs, SVMs, and RFs to predict daily market movements of Bitcoin, Ethereum, Ripple, and Litecoin. They compared subsets of feature vectors, including Twitter and market data, to identify the most effective models. Evaluation metrics such as accuracy, precision, recall, and F1 scores were employed. The results indicated that for Bitcoin, the MLP model incorporating both Twitter and market data performed the best, achieving an accuracy of 0.72 and a precision of 0.74, significantly outperforming random predictions. Ethereum's optimal model, an MLP utilizing both data sources, achieved an accuracy of 0.44 and a precision of 0.56, slightly better than random. Ripple's top-performing model was an MLP using only market data, with an accuracy of 0.64 and a precision of 0.68. In contrast, Litecoin's best model was an SVM utilizing both data sources, achieving an accuracy of 0.66 and a precision of 0.8. Notably, Twitter data alone was ineffective in predicting market movements. MLP consistently outperformed SVM and RF models across the studied cryptocurrencies. The research highlighted varying levels of predictability among the cryptocurrencies, with Litecoin demonstrating the highest precision, followed by Bitcoin and Ripple, while Ethereum showed lower predictability. Although Twitter data showed potential for Ripple and Litecoin, its effectiveness did not surpass that of market data alone. The study suggested that exploring diverse model approaches and understanding data integration nuances could enhance predictive capabilities further.

In contrast, Raju and Tarif (2020) focused specifically on Bitcoin price prediction and compared the performance of LSTM and ARIMA models. Results indicated that LSTM models, despite longer compilation times (61 milliseconds for LSTM vs. 4 milliseconds for ARIMA), showed promise in capturing the high fluctuations in Bitcoin's price data. Although ARIMA models had slightly lower Root Mean Square Error (RMSE) (209.263 for ARIMA vs. 198.448 for LSTM with single feature and 197.515 for LSTM with multi-feature), LSTM demonstrated greater suitability for Bitcoin price forecasting, especially in scenarios with significant price volatility. This suggests that LSTM models are better suited to handle the inherent unpredictability of Bitcoin prices, thus offering more accurate forecasting capabilities. Wolk (2020) took a different approach by employing various models, including support vector regression, stochastic gradient descent, gradient boosting model, multilayer perceptron neural network, least squares linear regression, AdaBoost, Bayesian ridge regression, decision tree, ElasticNet, and a hybrid model, to forecast cryptocurrency prices. Model performance was assessed using mean error (ME), correlation coefficient (R2), and actual error (T.s.). The hybrid model emerged as the most effective, exhibiting an error of less than $6 when predicting the closing price point across different cryptocurrencies. Furthermore, empirical trading experiments demonstrated that this approach generated a profit of $14.82 over one month, surpassing the performance of the KryptoBot tool, which yielded only $2.45 profit. Despite the inherent volatility in cryptocurrency markets, the proposed method proved to be profitable and stable, particularly during periods of market stability like Dec 2018. The study emphasizes the benefits of integrating sentiment analysis of Twitter data with machine learning models for cryptocurrency price prediction and trading.

On the other hand,Haritha and Sahana (2023) focused on the effectiveness of various sentiment prediction methods and the GRU model in cryptocurrency price prediction. FinBERT outperforms Bi-LSTM and GRU in sentiment prediction, achieving a lower mean absolute percentage error (MAPE) of 9.06% on average across two time periods, compared to 12.17% for Bi-LSTM and 11.20% for GRU. Regarding price prediction using GRU, the model demonstrates a low average MAPE of 3.6% for the period of 5 Aug 2022 to 5 Sept 2022, and 3.77% for 5 Sept 2022 to 5 Oct 2022. The comparison between real and predicted prices by the GRU model highlights its accuracy in capturing price trends. These results indicate that FinBERT offers superior sentiment prediction performance, while GRU exhibits robust capabilities in price prediction, making them suitable choices for their respective tasks.

Similarly,Singh and Bhat (2024) explored the effectiveness of transformer-based architectures in Ethereum price prediction. With a mean squared error (MSE) of 0.0051, the transformer model outperforms LSTM, achieving 2.59 times better results when using only Ethereum price data. Although LSTM exhibits better performance in terms of root mean squared error (RMSE) and mean absolute percentage error (MAPE), the transformer model surpasses artificial neural networks (ANN) and multilayer perceptron (MLP) models overall. Despite the dataset's smaller size and complexity compared to other studies, the transformer model demonstrates promising potential for Ethereum price prediction. Furthermore, integrating sentiment analysis data enhances the predictive capabilities of the transformer model. The comparison between predicted and actual Ethereum prices underscores the model's efficacy.

In terms of identifying market manipulations, Mirtaheri et al. (2021) conducted classification tasks to predict cryptocurrency market manipulations, emphasizing the effectiveness of Twitter features and economic features in different scenarios. The study presents findings from two classification tasks: Task I and Task II, focused on predicting cryptocurrency market manipulations. In Task I, utilizing a combination of Twitter-only, economic-only, and combined features, the average AUC score across all coins is 0.74, significantly outperforming the random baseline. Social media features, particularly from Twitter, prove more effective, although the addition of economic features slightly enhances performance. Task II, however, exhibits lower average accuracy compared to Task I, with economic features proving more useful (average AUC of 0.7) than Twitter features (average AUC of 0.59). Interestingly, combining Twitter features with economic ones deteriorates performance in Task II. Furthermore, the study identifies a correlation between prediction accuracy and coin features such as market cap and volume. Additionally, analysis reveals the prevalence of Twitter bots in cryptocurrency pump attempts, with an increasing ratio of bot activity observed among highly active users, emphasizing the role of bots in market manipulation.

Furthermore, Cheng et al. (2024) highlighted the comparable performance of the LTSM model to SARIMA and Fb-Prophet in predicting cryptocurrency prices, with LTSM excelling in forecasting volatility. Its strength lies in effectively leveraging past information for long-term predictions and dependencies. The analysis emphasized Bitcoin's high volatility, significantly, influenced by external factors like regulatory changes and social media activity. Recent geopolitical events such as the Russian-Ukrainian conflict and the COVID-19 pandemic have notably impacted cryptocurrency markets, with cryptocurrencies sometimes acting as safe havens or hedges. The study suggests that LTSM, SARIMA, and Fb-Prophet models are versatile for predicting Bitcoin values, with LTSM demonstrating superior performance based on MSE and MAE metrics. Specifically, LTSM yielded an MSE of 503,746.37 and an

MAE of 624.37, outperforming SARIMA (MSE: 38,080,003.02, MAE: 6170.9), Fb Prophet (MSE: 131,383,832.22, MAE: 7767.94), and GK volatility models (MSE: 1.719215e-07, MAE: 0.00011). Overall, the study offers valuable insights into understanding and forecasting cryptocurrency price movements amidst uncertain geopolitical and economic landscapes. El Haddaoui et al. (2023) The assessment criteria utilized in this research, namely Root Mean Squared Error (RMSE) and R-squared (R2), illustrate the effectiveness of the LSTM predictor in predicting Bitcoin prices based on specific attributes. The model attained an RMSE of 0.0368 and an R2 value of 0.974, indicating minimal prediction error and a strong correlation between forecasted and actual prices. Visual representations of market and social data further corroborate the model's efficacy by revealing associations between metrics like tweet volume, likes count, and price fluctuations. Training the model with combinations of social and market data significantly bolstered its performance, with the incorporation of all social and market features yielding optimal outcomes. Notably, metrics such as likes count and total positive tweets emerged as robust predictors of Bitcoin prices, emphasizing the influence of social engagement and sentiment on market dynamics. These findings offer valuable insights into comprehending and anticipating cryptocurrency price trends, underlining the necessity of integrating both social and market data for precise forecasting.

Moreover, Onyekwere, Ogwueleka, and Irhebhude (2022) examined the effectiveness of multilayer perceptron (MLP) models and autoregressive integrated moving average (ARIMA) models in forecasting cryptocurrency market trends in Nigeria. Visual representations of Bitcoin transaction volume over time revealed trends and fluctuations. MLP model results indicated varying levels of predictive accuracy, with Model 2 achieving the highest accuracy of 98.18%. In contrast, the ARIMA model's performance showed a decline in predictability over time, with a significant mean absolute percentage error (MAPE) of 46.13%. Comparative analysis revealed that MLP models outperformed ARIMA in terms of mean squared error (MSE), root mean squared error (RMSE), and MAPE. Discrepancies between forecasted values of ARIMA and MLP models were also evident. These results underscore the superior effectiveness of MLP models over ARIMA and their potential for accurate cryptocurrency market forecasting in Nigeria.

In another study, Uras and Ortu (2021) conducted a comparative analysis of restricted and unrestricted models to investigate blockchain cryptocurrencies' price movements using deep learning techniques. The evaluation relied on classification error metrics, including accuracy, F1 score, precision, and recall. In the restricted model, Support Vector Machines (SVM) achieved the highest average accuracy of 54%, followed by neural networks with Convolutional Neural Networks (CNN) leading at 53.7% accuracy. However, in the unrestricted model, neural networks, especially CNN, surpassed other models with a mean accuracy of 54.7%. Moreover, incorporating technical analysis variables in the unrestricted model contributed to improved average accuracy across all implemented models, indicating its efficacy in enhancing prediction accuracy. The study acknowledges potential threats to validity, such as limited generalizability due to analyzing only one cryptocurrency and unaccounted confounding factors influencing price movements. Nonetheless, the research offers valuable insights into the performance of various models in predicting cryptocurrency price trends. Othman et al. (2020) conducted a study focusing on improving the prediction accuracy of Bitcoin market prices through an artificial neural network (ANN) approach implemented using RapidMiner software. Descriptive statistics revealed significant volatility in Bitcoin prices, with mean values for various metrics such as BTC PRICE, Open, High, Low, and Close. These metrics exhibited corresponding standard deviations, indicating substantial fluctuations. Following data normalization, the distributions became normal, facilitating further analysis. The neural network model achieved an impressive prediction accuracy of 92.15% for the Bitcoin Price Index, with the volatility of the low price index predicting daily Bitcoin prices with 63% accuracy. Regression plots generated by the ANN model demonstrated an optimal fit, with correlation values reaching 99.96%, highlighting its effectiveness in forecasting Bitcoin prices. The study emphasized the importance of asymmetric information in accurately predicting Bitcoin market prices, attributing significance to supply and demand dynamics rather than macroeconomic factors or corporate news. Jagannath et al. (2021) introduced a self-adaptive technique, the jSO optimization algorithm, for developing a deep learning model aimed at predicting Bitcoin prices. The study's results highlight the effectiveness of this approach, as the algorithm efficiently identifies optimal model parameters, resulting in predictions that closely match actual Bitcoin prices with high accuracy. A graphical representation comparing predicted and actual Bitcoin prices illustrates the model's ability to accurately track price fluctuations over time. Furthermore, a mean absolute error (MAE) analysis reveals that the jSO-based LSTM model consistently outperforms traditional LSTM models across various training sizes, with MAE improvements ranging from 32.9% to 49.2%. These findings underscore the superiority of the jSO-based LSTM model in generating more precise and timely predictions of Bitcoin prices, indicating its potential as a valuable tool for cryptocurrency price forecasting. Lamothe-Fern´andez et al. (2020) offered a thorough examination of Bitcoin price prediction using deep learning techniques, yielding promising outcomes: the models achieve prominent levels of accuracy ranging from 92.61% to 97.34%. Descriptive statistics reveal an initial stability in the data, with significant variability observed in key variables such as transaction value, volume, block size, and difficulty. Notably, variables like block size, cost per transaction, and difficulty consistently demonstrate significance across all models, while others like the dollar exchange rate and Dow Jones index exhibit importance in specific models. Post-estimation analysis confirms the reliability of the models, with the Deep Recurrent Convolution Neural Network (DRCNN) consistently performing better than others. These findings provide valuable insights into Bitcoin price dynamics, emphasizing the importance of both demand-supply dynamics and macroeconomic indicators' inaccurate price prediction.

Additionally, Ghosh, Jana, and D. K. Sharma (2024) proposed a novel predictive modeling framework that exhibits exceptional effectiveness, outperforming other models in both normal and pandemic periods. Utilizing the Diebold-Mariano (DM) test, its statistically significant performance superiority is evident, especially in forecasting Bitcoin (BTC) and Tether (USDT) prices during the COVID-19 pandemic. BTC stands out as the most predictable cryptocurrency, with an average Directional Accuracy (DA) of 87.5%, whereas USDT shows the lowest predictability, with an average DA of 42.9%. With its ability to accurately predict trend movements, the framework assists traders in making profitable investments and managing risks. Its scalability allows for precise monitoring of market dynamics, enabling short-term financial planning even in crisis situations. Additionally, the framework's superiority over various established models in different scenarios highlights its reliability. Serving as a robust business analytics tool, it provides actionable insights for enhancing business growth and sustainability. Overall, the integrated forecasting model comprising MODWTXGB, EEMED-LSTM, and NLP demonstrates statistical superiority over RNN, GRNN, RF, and QRRF models, enriching business analytics research with its practical implications and statistical robustness. Ortu et al. (2022) The analysis of cryptocurrency price classification tasks reveals significant insights. Combining trading and social media indicators notably boosts classification accuracy, with the unrestricted model achieving the best performance, particularly with the CNN architecture, reaching an average accuracy of 87% and a standard deviation of 2.7%. In contrast, the restricted model shows accuracy ranging from 51% to 84% across various algorithms, highlighting the advantages of the unrestricted approach. Technical indicators alone generally outperform models incorporating social and trading indicators, especially in daily frequency classification. Ethereum's price movements demonstrate higher accuracy, precision, recall, and F1 scores compared to Bitcoin. Moreover, during periods of financial turmoil in cryptocurrency markets, integrating social media indicators notably enhances F1 scores, as confirmed by the Wilcoxon Signed Rank Test results, rejecting the null hypothesis at a 5% significance level. These findings emphasize the importance of incorporating diverse indicators, particularly social media, for precise prediction of cryptocurrency price movements, particularly amid market volatility. Ibrahim, Kashef, and Corrigan (2021) revealed that ARIMA achieved an accuracy of 51.77%, Prophet attained 52.60% accuracy, and Random Forest models achieved accuracies of 50.51% and 50.89%, respectively. The MLP model demonstrated the best performance among the models with an accuracy of 54.09%, although it was not higher than the momentum strategy's accuracy of 53.85%.

Furthermore, Jaquart, Dann, and Weinhardt (2021) examined predictive models showing that all models achieve predictive accuracy surpassing 50%, indicating substantial accuracy beyond random chance. Increased stability is observed with multiple random seeds, particularly for stochastic models. Among the models, LSTM and GBC consistently emerge as top performers, with accuracy notably improving for longer prediction horizons. Diebold-Mariano tests reveal significant differences in accuracy between models. Feature importance analysis highlights the decreasing significance of minutely bitcoin return time series for RNNs over longer horizons, while other features become increasingly important. Although a trading strategy based on model predictions generates positive returns before transaction costs, transaction costs notably impact trading performance, leading to negative returns for all methods.

Lastly, Carbó and Gorjón (2024) conducted an analysis of Bitcoin price determinants using machine learning and interpretability techniques, focusing on the LSTM model's performance evaluated by RMSE across three distinct periods: launch, consolidation, and expansion. In the launch period, RMSE ranged from 5.3% to 5.7%, showing reasonable accuracy, while in the consolidation phase, it increased to 13.2% to 21.2%. During the expansion period, RMSE ranged from 15.1% to 19.1%. Notably, employing the rolling window approach yielded results akin to long-range forecasts, indicating robustness in the model. The SHAP analysis underscored the importance of technological variables in the early periods, gradually transitioning to public interest variables. In the launch period, technological features contributed 50% to the model's explanatory power, declining to 21% in 2021, while interest variables increased from 11% to 34% over the same period. The observed performance variations between periods suggest potential influences of latent factors such as lead-lag relationships and speculative bubbles on Bitcoin price dynamics. Nguyen, Crane, and Bezbradica (2022) explored the relationship between Twitter-based sentiment indicators (from ChatGPT, BERT, and VADER) and Bitcoin returns. ChatGPT and BERT exhibited synchronized sentiment movements, particularly from mid-2018 to mid-2020, while VADER did not align as closely. Regression analysis revealed that only ChatGPT's sentiments (Bullish Index, Variation, and Agreement) significantly predict daily excess Bitcoin returns. Further robustness checks and comparative analysis using a horse race model confirmed that ChatGPT's sentiments outperform both BERT and VADER, as well as traditional sentiment proxies, in predicting Bitcoin returns. These results underscore ChatGPT's superior ability to capture market sentiment nuances missed by other models.

## 5.2 Sentiment Analysis Models

### 5.2.1 Traditional Machine Learning Approaches

Sentiment analysis, a crucial aspect of understanding market dynamics, has been extensively explored in the context of cryptocurrency using traditional machine-learning techniques. Several studies have contributed to this area, employing various algorithms and data sources to analyze sentiment from social media platforms like Twitter, Reddit, and YouTube. This section of the chapter reviews and compares these studies' findings to provide insights into the effectiveness of different approaches. Dulău and Dulău (2019) investigated sentiment analysis of social media content related to cryptocurrencies, utilizing platforms like Coin Market Cap, Twitter, and Reddit.

**Table 2: Performance Evaluation of Price Prediction Models**

| Papers | Subcategories | Models | Metrics | Best Model | Expected Result |
|---|---|---|---|---|---|
| (Greaves and Au 2015) | Regression, Classification | Linear Regression, Logistic Regression, and SVM | MSE, Classification Accuracy | Neural Network | 55.10% |
| (Rahman et al. 2018) | Regression, Classification | MLR, SVR, RF, and NB | Accuracies of various models | Naive Bayes | 89.65% |
| (Jung et al. 2023) | Classification | XGBoost, LightGBM, RF, NB, and SVM | Accuracy, AUC | XGBoost | 90.57% |
| (Abraham et al. 2018) | Regression | Linear Regression | Correlation metrics, Non-linear trends | | |
| (Lamon, Nielsen, and Redondo 2017) | Classification | Logistic Regression, Linear SVM, Multinomial NB, and Bernoulli NB | Accuracy in predicting price increases/decreases | Logistic Regression and BNB | Bitcoin: 61.9% (Price decrease) Ethereum: 75.8% (price decrease) Litecoin: 100% (price decrease) |
| (Zhu et al. 2023) | Regression | SVR, LSSVR, TWSVR, and ARIMA | EVS, R2, MAE, MSE, RMSE, MAPE, CPU running time | XGBOOST-WOA -TWSVR | EVS-0.9547 |
| (Akyildirim, Goncu, and Sensoy 2021) | Classification | Logistic Regression, SVM and ANN | in-sample accuracies | ANN | 97% |
| (Bâra and Oprea 2024) | Regression | Extreme Learning Machine and XGBR | MAE, RMSE | Stacked Model | MAE: 595.743 |
| (Valencia, Gómez-Espinosa, and Valdés-Aguirre2019) | Deep Learning | MLPs, SVMs and RFs | Accuracy, Precision, Recall, F1 scores | MLP | Bitcoin: 72% Ethereum: 44% Ripple: 64% |
| (Raju and Tarif 2020) | Deep Learning | LSTM and ARIMA | RMSE | LSTM | RMSE: 197.515 |
| (Wolk 2020) | Deep Learning | MLP and Hybrid Model | ME, R2, Actual error, Profit | Hybrid Model | ME: 13.79 |
| (Haritha and Sahana 2023) | Deep Learning | FinBERT, GRU and BiLSTM | MAPE | GRU | |
| (Singh and Bhat 2024) | Deep Learning | Transformer based NN, ANN, and MLP | MSE, RMSE, MAPE | Transformer + ETH data | MSE:0.0051 |
| (Mirtaheri et al. 2021) | Classification | SVM | AUC | | 87% |
| (Cheng et al. 2024) | Deep Learning | LTSM, SARIMA and Fb-Prophet | MSE, MAE | LSTM-GK Volatility | MAE: 0.00011 |
| (El Haddaoui et al. 2023) | Deep Learning | LSTM | RMSE, R2 | | |
| (Onyekwere, Ogwueleka, and Irhebhude 2022) | Deep Learning | MLP and ARIMA | MAPE, MSE, RMSE | MLP | 98.18% |
| (Uras and Ortu 2021) | Deep Learning | SVM and CNN | Accuracy | CNN | 54.70% |
| (Othman et al. 2020) | Deep Learning | ANN | Prediction accuracy | ANN | 92.15% |
| (Jagannath et al. 2021) | Deep Learning | JSO based LSTM and LSTM | MAE | jSO based LSTM | MAE: 1.89352 |
| (Lamothe-Fern´andez et al. 2020) | Deep Learning | Deep Recurrent Convolution Neural Network (DRCNN), Deep Neural Decision Trees (DNDT), and Deep Learning Linear Support Vector Machines (DSVR) | Accuracy | DRCNN | 97.34% |
| (Ghosh, Jana, and D. K. Sharma 2024) | Deep Learning | Maximal Overlap Discrete Wavelet (MODWT) and the Ensemble Empirical Mode Decomposition (EEMD), XGBoost (XGB) and LSTM | Directional Accuracy | MODWT-XGB, EEMED-LSTM and NLP | Bitcoin: 87.5% |
| (Ortu et al. 2022) | Deep Learning | MLP, LSTM, MALSTM-FCN, and CNN | Accuracy | CNN | Bitcoin-83% Ethereum: 84% |
| (Ibrahim, Kashef, and Corrigan 2021) | Deep Learning | ARIMA, Prophet, RF and MLP | Accuracy | MLP | 54.09% |
| (Jaquart, Dann, and Weinhardt 2021) | Deep Learning | LSTM, GRU, RF, and GBC | Predictive accuracy | LSTM | 56% |
| (Carbó and Gorjón2024) | Deep Learning | LSTM | RMSE | LSTM (launch period) | RMSE: 5.3% |
| (Mahdi et al. 2021) | Classification | SVM, linear kernel, polynomial kernel, and radial kernel | Accuracy | SVM- radial kernel | Bitcoin: 91.8% |
| (Nguyen, Crane, and Bezbradica2022) | Deep Learning | AT-LSTM-MLP,LSTM,SVR, RF and TCN | RMSE, MAE and SMAPE | AT-LSTM-MLP | MAE: 1.53 |

The study employs language processing techniques such as Stanford CoreNLP and IBM Watson, adapting algorithms to accommodate cryptocurrency-specific jargon that influences sentiment. Results show enhanced accuracy compared to conventional approaches, with consistent outcomes across various data sources, indicating low sensitivity to verification levels.

Similarly, Aslam et al. (2022) proposed an approach for sentiment analysis and emotion detection in cryptocurrency-related tweets, achieving notable results across various evaluation metrics. SVM and LR machine learning models demonstrated high accuracy scores of 0.90 using Bag-of-Words (BoW) and TF-IDF features for emotion detection. In a related study, Prasad, Sharma, and Vishwakarma (2022) displayed the effectiveness of various machine learning algorithms in analyzing sentiment from YouTube comments related to cryptocurrency. Accuracy rates ranged from 89.1% to 94%. Logistic Regression achieved the highest accuracy of 93.6%, followed by Decision Tree Classifier with 89.5%. Support Vector Machine and Gaussian Naive Bayes attained accuracies of 93% and 90.9%, respectively. Ensemble techniques like Random Forest and Extra Trees Classifier yielded accuracies of 92.4% and 93.1%, while AdaBoost and Gradient Boosting Classifier scored 90% and 92%. Impressively, the stacking ensemble method surpassed all others, achieving a remarkable accuracy of 94%. These results underscore the effectiveness of ensemble learning, particularly stacking, in sentiment analysis tasks involving YouTube comments on cryptocurrency.

Additionally, Jain et al. (2018) conducted analysis on Bitcoin and Litecoin data and revealed notable insights. For Bitcoin, the multiple linear regression (MLR) models accounted for 44% of the variation in its price, yet none of the variables—positive tweets, neutral tweets, or negative tweets—proved statistically significant based on their p-values. Additionally, the Durbin-Watson statistic indicated positive autocorrelation in the residuals. Conversely, for Litecoin, the MLR model performed better, explaining 59% of its price variation. In this case, the coefficients for positive and neutral tweets, along with negative tweets, were statistically significant. Both Bitcoin and Litecoin exhibited positive autocorrelation in their residuals. Furthermore, analysis of tagged tweets highlighted that positive and neutral tweets had a more substantial impact on cryptocurrency prices compared to negative tweets, reflecting their higher frequency and consequent influence on price fluctuations.

### 5.2.2 Deep Learning Models

In recent years, deep learning models have emerged as powerful tools for sentiment analysis and emotion detection in cryptocurrency-related social media content. Numerous studies have proposed innovative approaches and compared their performance to traditional machine learning methods. Aslam et al. (2022) proposed a deep-learning approach for sentiment analysis and emotion detection in cryptocurrency-related tweets. They developed an LSTM-GRU ensemble model, which achieved an impressive accuracy of 0.99 for sentiment analysis. When applied to a balanced dataset using random undersampling, the accuracy slightly declined to 0.97 for sentiment analysis and 0.83 for emotion detection.

Moreover, the LSTM-GRU model exhibited faster computational times, completing sentiment analysis in just 1178 seconds (about 39 minutes) and emotion detection in 1438 seconds (about 48 minutes), outperforming other models in terms of efficiency. The proposed LSTM-GRU model outperformed state-of-the-art approaches, highlighting its effectiveness in sentiment analysis and emotion detection tasks. Similarly, Habek, Toçoğlu, and Onan (2022) experimented with different deep neural network architectures for cryptocurrency sentiment analysis. They observed that their proposed CGWELSTM architecture consistently outperformed conventional methods, achieving an average accuracy of 93.77% and demonstrating improved precision, recall, and F-measure values. Furthermore, the CGWELSTM-v3, incorporating bi-LSTM in the bidirectional layer, achieved the second-highest accuracy at 90.97%. Their architecture also surpassed existing models on the Sentiment dataset, achieving an accuracy of 85.40%. Furthermore, Huang et al. (2021) compared LSTM-based sentiment analysis with an auto-regression approach. The evaluation focused on analyzing sentiment extracted from Sina-Weibo posts authored by the top 100 cryptocurrency investors. The results indicated that the LSTM model achieved a precision rate of 87.0% and a recall rate of 92.5%, outperforming the AR method by 18.5% in precision and 15.4% in recall. These outcomes underscore the superior capability of LSTM in accurately gauging sentiment from social media content, particularly in the context of cryptocurrency investment trends. Moudhich and Fennan (2024) explored graph embedding techniques for sentiment analysis in cryptocurrencies and found them to outperform word embedding and simple embedding models. The graph embedding model exhibited the highest accuracy, reaching 0.91, surpassing the word embedding (0.87) and simple embedding (0.82) models. Moreover, the graph embedding model demonstrated robust performance across various metrics, including precision (0.91 and 0.90), recall (0.90 and 0.89), and overall accuracy (0.92). In terms of sentiment analysis of the cryptocurrency market, the graph embedding model identified 69.30% positive sentiment, while the word embedding and simple embedding models showed 55.41% and 51.16% positive sentiment, respectively. Additionally, Dwivedi (2023) developed a BERT-based sentiment analysis model with high accuracy and validated its effectiveness in accurately classifying sentiment in cryptocurrency-related social media posts. The model achieved a remarkable accuracy of 98.23% after 4 training epochs and maintained a high overall accuracy of 89% across 11,075 samples. The precision, recall and F1 score metrics vary for different sentiment categories, with Positive sentiment being the most prevalent, followed by Negative and Neutral sentiments. Additionally, the model demonstrates effectiveness with mean absolute and mean squared errors of 0.2173 and 0.4267, respectively. Kulakowski and Frasincar (2023) also evaluated various sentiment classification models for cryptocurrency-related social media posts on both StockTwits and emoji test sets and found BERT-based models to outperform others, highlighting the effectiveness of deep learning models in sentiment analysis tasks. The CryptoBERT XL model achieves the highest performance on the StockTwits test set, with an accuracy of 58.49% and a macro F1 score of 58.83%. LUKE, on the other hand, outperforms the VADER lexicon on the emoji test set. The study by Şaşmaz and Tek (2021) conducted sentiment analysis on cryptocurrencies using the Random Forest classifier, achieving promising results with 82% accuracy for the training set and 77% for the test set. Notably, optimized parameters included 200 estimators, 'log2' for max features, and 'gini' for criterion. However, compared to a pre-trained BERT model, the Random Forest classifier significantly outperformed it, achieving a higher accuracy of 77% compared to BERT's 45%. BERT's accuracy increased to 53% when considering only positive and negative classes. The research found that neutral sentiments had the highest correlation (45%) with NEO price according to the Random Forest model, while BERT sentiments showed the highest correlation (46%) with positive sentiments. Furthermore, correlations were observed between NEO price and ETH (67%) and BTC (41%) prices, alongside a strong correlation of 91% between BTC and ETH prices.

Furthermore, Wong (2021) compared the performance of Naïve Bayes and LSTM models using both synthetic and real data. In the synthetic data analysis, Naïve Bayes outperformed LSTM in distinguishing similar tweets based on sentiment, correctly identifying 6 out of 7 examples compared to LSTM's 4 out of 7. However, neither model could establish a consistent threshold for categorizing tweets as positive or negative sentiment. In terms of quantitative evaluation, Naïve Bayes showed random performance with 50% accuracy, while LSTM exhibited slightly better performance with 36% accuracy on synthetic data. When applied to real data, LSTM displayed enhanced accuracy at 51%, showing a marginal improvement over Naïve Bayes. Kang, Hwang, and Shin (2024) leveraged ChatGPT-3.5 and ChatGPT-4.0 models for cryptocurrency sentiment analysis. With ChatGPT-3.5, daily sentiment scores averaged positive values globally, with an average sentiment score of 0.279 for headlines and bodies (HB) in global news articles and 0.108 for Korean HB. Correlation coefficients above 0.6 were observed between sentiment scores calculated using news article headlines, bodies, and both, while correlations between global and Korean news sentiments were around 0.3. Korean news sentiment led global sentiment, indicating predictive power. However, sentiment analysis did not significantly explain cryptocurrency factors or predict future cryptocurrency returns using ChatGPT-3.5. Nonetheless, Korean news sentiment could predict both Korean and global cryptocurrency prices, suggesting relative market inefficiency. ChatGPT-4.0 showed improved short-term predictability of global news sentiment compared to ChatGPT-3.5, but the predictive power of sentiment for cryptocurrency returns remained limited with both models. Miah et al. (2024) examined sentiment analysis for Arabic, Chinese, French, and Italian sentences translated into English using Libre Translate and Google Translate. We evaluated these translations with three pre-trained sentiment analysis models: Twitter-RoBERTa-Base, BERTweet-Base, and GPT-3, as well as a proposed ensemble model. Conducting 32 experiments, we found that the combination of Google Translate and the Proposed Ensemble model achieved the highest metrics across various performance indicators: accuracy (0.8671), precision (0.8091), recall (0.8122), F1 score (0.8106), and specificity (0.5713). These results underscore the efficacy of translating foreign language sentences for sentiment analysis, with GPT-3 and the ensemble model showing particularly impressive performance, especially for Chinese sentences. The proposed ensemble models outperformed established multilingual models like XLM-T and mBERT, suggesting that integrating machine translation with robust sentiment analysis models can effectively address multilingual sentiment analysis challenges.

## 5.3 Hybrid Models

The studies reviewed in this section collectively highlight the effectiveness of hybrid models, integrating sentiment analysis with machine learning techniques, in predicting cryptocurrency prices. Pant et al. (2018) emphasized the significance of sentiment analysis in forecasting cryptocurrency prices, demonstrating the superiority of a voting classifier in sentiment classification and its correlation with Bitcoin price fluctuations. The evaluation of methods for extracting features for sentiment classification shows that Bag-of-Words performs better than Word2Vector, achieving accuracies ranging from 77.95% to 78.49%, compared to 68.33% to 69.82%.

**Table 3: Performance Evaluation of Sentiment Analysis Models**

| Paper | Subcategories | Models | Metrics | Best Model | Expected Result |
|---|---|---|---|---|---|
| (Prasad, Sharma and Vishwakarma2022) | Traditional ML | Logistic Regression, DT, SVM, GNB,RF, Extra Trees, AdaBoost, GB and Stacking | Accuracy | Stacking ensemble method | 94% |
| (Jain et al. 2018) | Traditional ML | MLR | R^2, P-values, Durbin-Watson statistic | N/A | 44% variation explained for Bitcoin, 59% for Litecoin |
| (Aslam et al. 2022) | Deep Learning | LSTM-GRU Ensemble | Accuracy | LSTM-GRU Ensemble | 99% |
| (Habek, Toçoğlu, and Onan 2022) | Deep Learning | CGWELSTM, CGWELSTM-v3 | Accuracy | CGWELSTM | 93.77% |
| (Huang et al. 2021) | Deep Learning | LSTM | Precision, Recall | LSTM | Precision:87% |
| (Moudhich and Fennan 2024) | Deep Learning | Graph Embedding | Accuracy, Precision, Recall | Graph Embedding | 91% |
| (Dwivedi 2023) | Deep Learning | BERT | Accuracy | BERT | 98.23% |
| (Kulakowski and Frasincar 2023) | Deep Learning | BERT-based models | Accuracy | CryptoBERT XL | 58.49% |
| (Şaşmaz and Tek 2021) | Deep Learning | Random Forest and BERT | Accuracy | Random Forest | 77% |
| (Wong 2021) | Deep Learning | Naïve Bayes and LSTM | Accuracy | LSTM | 51% |
| (Pant et al. 2018) | Hybrid | Voting Classifier, RNN,GRU | Accuracy | Voting Classifier | 81.39% |
| (Inamdar et al. 2019) | Hybrid | Random Forest | MAE, RMSE | - | MAE: 2.7526 |
| (Wu 2023) | Hybrid | OLS, RF and LSTM | RMSE, MAE | RF and LSTM | Binance: RF: MAE-0.036 Bitcoin: RF: MAE: 3.328 Solana: LSTM: MAE: 5.68 |
| (Bao 2022) | Hybrid | VADER, LSTM, Bi-LSTM | Accuracy | Bi-LSTM | 98% |
| (Gurrib and Kamalov 2022) | Hybrid | SVM, LDA | Accuracy, Precision, Recall | SVM | 58.50% |
| (Parekh et al. 2022) | Hybrid | DL-GuesS | MSE, MAE, MAPE | DL-GuesS | Dash: MAE 0.0805 |
| (Yasir et al. 2023) | Hybrid | DL, SVR | MAE | SVR | Bitcoin: 27.451 |
| (Vo, Nguyen, and Ock 2019) | Hybrid | SVM and LSTM | MANE | LSTM | MANE: 1.36% |
| (Wimalagunaratne and Poravi 2018) | Hybrid | Neural Network | Accuracy | - | Ethereum: 85% |
| (Moudhich and Fennan 2024) | Graph Embedding | Graph embedding and Bi-LSTM | Accuracy | Model with Graph Embedding | 91% |
| (Wahidur et al. 2024) | Deep Learning | DistilBERT, MiniLM and FLAN-T5-Base | Accuracy, F1-Score, Precision, Recall | FLAN-T5-Base IT model (Instruction Tuning) | Average accuracy: 75.17% across all datasets. |
| (El Abaji and A Haraty 2024) | Hybrid | LSTM, Prophet, SARIMAX and LLaMA-2 | Accuracy | LSTM-with sentiment | 87.25% |
| (Kang, Hwang, and Shin 2024) | Deep learning | ChatGPT 4.0,<br><br>sBERT, GPT-3.5-<br><br>turbo-16k<br><br>and VAR | Excess Returns | ChatGPT-4.0 | 16-26 bps for Korean market portfolio |
| (Zuo, Chen, and Härdle 2024) | Hybrid | GPT-4 and FinBERT | Regression Coefficient | N/A | 5.42E+04 |

A voting classifier achieves an overall sentiment classification accuracy of 81.39%, with strong precision, recall, and F-measure scores. Examination of the Pearson Correlation Coefficient between sentiment scores and percentage price changes of Bitcoin reveals moderate correlations for negative sentiment (0.34 to 0.41) and weaker correlations for positive sentiment (0.21 to 0.26) for price fluctuations exceeding 2% and 4%. Additionally, the accuracy of price prediction using an RNN model is found to be 77.62%, supported by a comparative plot illustrating the model's performance against actual Bitcoin prices. These results emphasize the importance of sentiment analysis in forecasting cryptocurrency prices and offer insights into the effectiveness of different models and features.

Contrastingly, Inamdar et al. (2019) conducted experiments to assess the Random Forest model's effectiveness in predicting Bitcoin prices over the next two days using historical data and sentiment scores from news and Twitter. The results indicate relatively accurate predictions for the first day, with a Mean Absolute Error (MAE) of 2.7526 and a Root Mean Square Error (RMSE) of 13.7033. However, the accuracy slightly decreases for the second day, with an MAE of 3.1885 and an RMSE of 15.1686. The analysis suggests that social media sentiment scores have minimal impact on price prediction, due to their neutral nature. Additionally, Wu (2023) shed light on the performance of machine learning models in predicting daily cryptocurrency price returns, focusing on Bitcoin, Solana, and Binance, with and without incorporating sentiment data from social media. Initially, without sentiment data, models struggled to accurately predict Bitcoin's daily price returns, with an average error of 4.5-4.7% RMSE and 3.3-3.5% MAE. Solana's predictions were even less accurate, with RMSE ranging from 7.411 to 9.777 and MAE ranging from 5.680 to 7.873. In contrast, Binance showed greater accuracy due to its stability, with RMSE ranging from 0.041 to 0.200 and MAE ranging from 0.031 to 0.159. However, after incorporating sentiment scores related to Bitcoin, marginal improvements were observed in Bitcoin's prediction accuracy, while significant improvements were noted for Solana, reducing its RMSE to 7.411 and 5.680 and MAE to 5.680 and
7.873, respectively.

Conversely, for Binance, adding sentiment data led to increased noise in the model, resulting in decreased accuracy, with RMSE ranging from 0.041 to 0.046 and MAE ranging from 0.031 to 0.036. Notably, machine learning models that did not assume linearity performed better with sentiment data, indicating the complexity of incorporating sentiment into cryptocurrency price prediction models. Bao (2022) delved into the relationship between sentiment analysis and Bitcoin prices through the application of Natural Language Processing (NLP) techniques. It introduced two models: one utilizing Vader sentiment for sentiment analysis and the other employing neural network algorithms like LSTM and Bi-directional LSTM. While the first model achieved an accuracy of approximately 60%, the second model, based on Bi-directional LSTM, significantly outperformed it with an accuracy of 98%. Moreover, comparisons were made between different data input methods, revealing a strong correlation between predicted and observed Bitcoin prices through imputation results. Additionally, uni-variable, and multi-variable experiments were conducted, with the latter incorporating sentiment scores alongside time as independent variables.

Although the multi-variable experiments resulted in a 2.3x increase in running time, they yielded higher accuracy and reduced percent error. In conclusion, the study highlights the effectiveness of combining NLP with tspDB for predicting future Bitcoin prices based on sentiment analysis of textual data. Gurrib and Kamalov (2022) took a different approach by aiming to predict Bitcoin (BTC) price movements by integrating sentiment analysis with asset-specific data. Analyzing five years of data (2016–2021), the study compared various models, including random guess predictions, models without sentiment analysis, and SVM models from similar studies. Evaluation metrics such as precision, recall, and accuracy were used to assess model performance. Results showed that the proposed approach surpassed random guess predictions, achieving accuracy values between 0.525 and 0.585. SVM models incorporating news sentiment and asset-specific information demonstrated the highest accuracy (0.585), with significant improvements in forecasting both upward and downward movements in BTC prices. Conversely, LDA models exhibited varied performance, with asset-specific information alone achieving the best precision score (0.667) for predicting downward movements. The study highlights the effectiveness of the proposed method in outperforming random guess approaches and achieving accuracy levels comparable to or higher than those reported in recent literature. These findings support previous studies favoring SVM models for BTC price prediction and emphasize the importance of sentiment analysis combined with asset-specific information in enhancing BTC price forecasting accuracy.

Moreover, Parekh et al. (2022) evaluated DL-GuesS, a cryptocurrency price prediction model based on deep learning and sentiment analysis, against traditional models. DL-GuesS, trained using TensorFlow APIs and sentiment analysis from Twitter data with the VADER algorithm, outperformed traditional models in predicting Dash and Bitcoin-Cash prices, achieving lower mean squared error (MSE), mean absolute error (MAE), and mean absolute percentage error (MAPE) values. For Dash price prediction, DL-GuesS attained an MSE of 0.0185, MAE of 0.0805, and MAPE of 4.7928, while for Bitcoin-Cash, it achieved an MSE of 0.0011, MAE of 0.0196, and MAPE of 4.4089. These findings highlight DL-GuesS's effectiveness in cryptocurrency price prediction, suggesting its potential application in real-world trading scenarios. Similarly, Yasir et al. (2023) reveals the efficacy of different regression models in predicting cryptocurrency prices, with Deep Learning (DL) and Support Vector Regression (SVR) outperforming Linear Regression (LR) in most scenarios. SVR demonstrates superior accuracy over LR across the top five cryptocurrencies, while DL shows the best overall performance. Specifically, for Bitcoin, Dash, Litecoin, Monero, and Stellar, SVR achieves Mean Absolute Error (MAE) values of 27.451, 2.600, 0.731, 1.150, and 0.002 respectively, whereas DL achieves MAE values of 41.865, 4.399, 1.475, 2.038, and 0.003, respectively. The inclusion of sentiment analysis further enhances prediction accuracy, particularly for DL, which consistently outperforms SVR and LR across various cryptocurrencies and sentiment inputs. For instance, considering sentiments related to events like Brexit (2016), Refugees Welcome (2015), the Gaza Attack (2014), and the Hong Kong protest (2014), DL achieves improved MAE values compared to SVR and LR for cryptocurrencies such as Bitcoin, Litecoin, Dash, and Monero.

Additionally, Vo, Nguyen, and Ock (2019) conducted a study assessing a model for Ethereum (ETH) price prediction using daily time series data. The model integrated sentiment analysis of news over the past 7 days along with historical price data from the previous 7 days. Experimental findings showed that the model's predictions closely matched actual ETH prices, with the weakest performance observed during significant price fluctuations. Performance evaluation, measured by mean absolute normalized error (MANE), indicated satisfactory outcomes for the proposed model compared to alternative methods. Notably, the LSTM model, particularly when incorporating sentiment analysis features (MANE=1.36%), outperformed support vector regression (SVM) (MANE=3.52%), highlighting the efficacy of LSTM networks for time series data prediction when augmented with sentiment analysis. El Abaji and A Haraty (2024) investigated the effect of integrating sentiment analysis, using LLaMA-2, on the predictive performance of Bitcoin price forecasting models, specifically LSTM, Prophet, and SARIMAX. Initially, without sentiment analysis, the LSTM model demonstrated a MAPE of 15.59% with an accuracy of 84.41%, the Prophet model showed a MAPE of 13.73% and an accuracy of 86.27%, while the SARIMAX model achieved a MAPE of 13.28% with an accuracy of 86.72%. The inclusion of sentiment analysis significantly enhanced the performance of the LSTM model, reducing the MAPE to 12.75% and improving accuracy to 87.25%, thus providing a closer alignment with actual price movements.

In contrast, the Prophet and SARIMAX models displayed mixed outcomes. For the Prophet model, the MAPE increased to 28.04% and accuracy decreased to 71.96%. The SARIMAX model's metrics deteriorated markedly, underscoring the difficulties in effectively incorporating sentiment data for these models. Zuo, Chen, and Härdle (2024) evaluated emoji sentiment analysis in predicting Bitcoin (BTC) prices. Positive emoji sentiments, particularly those derived from the top 5 and top 10 tweets, exhibit a strong predictive relationship with the subsequent day's BTC prices. Regression analysis underscores this with coefficients of 5.322e+04 and 5.417e+04, respectively, indicating a robust positive correlation. Conversely, the bottom 5 and bottom 10 tweets' sentiments do not present significant correlations, highlighting the pronounced impact of positive sentiment on market movements. This asymmetry suggests that optimistic social media discourse, as encapsulated by emojis, plays a crucial role in forecasting market trends.

Lastly, Wimalagunaratne and Poravi (2018) assessed the cryptocurrency prediction system involved in examining the correlation between tweet polarity and Bitcoin price changes, revealing varied correlations and a delayed market response to online discussions. Despite this, the system's neural network capabilities helped offset any accuracy issues. Testing on three main cryptocurrencies over a three-month period indicated Ethereum had the highest accuracy at 85%, followed by Bitcoin (exact accuracy unspecified), and Bitcoin Cash with the lowest accuracy of 70%. Ethereum's higher accuracy may be attributed to its incorporation of Bitcoin's predicted price changes into the model, while Bitcoin Cash's lower accuracy reflects its smaller dataset. These results highlight the system's effectiveness in less volatile market conditions and its adaptability to different cryptocurrencies with varying levels of accuracy.

## 5.4 Market Dynamics and External factors

The literature on cryptocurrency price forecasting and volatility analysis encompasses a broad spectrum of methodologies and findings, reflecting the dynamic nature of digital asset markets and the evolving landscape of predictive analytics. Several studies have focused on understanding the intricate relationships between cryptocurrency prices and various external factors, such as economic indicators, market sentiment, and geopolitical events. Mahdi et al. (2021) and Basher and Sadorsky (2022) investigated the impact of macroeconomic factors, including daily COVID-19 dynamics and gold prices, on cryptocurrency returns and volatility. Mahdi et al. (2021) conducted empirical analysis utilizing real-world data to investigate the relationship between several factors, including daily confirmed COVID-19 infections and deaths, daily gold prices, and daily closing prices of major cryptocurrencies. The study observed a range of daily COVID-19 deaths and infections, with high volatility during the pandemic period.Similarly, daily gold prices exhibited a correlated trend with COVID-19 dynamics, displaying increased volatility during the pandemic compared to pre-pandemic times. Analyzing major cryptocurrencies' prices based on quartiles of gold prices, the study noted significant volatility spikes during the pandemic. Granger causality tests indicated a significant causality between gold prices and cryptocurrency returns, with certain cryptocurrencies influencing gold prices. The study emphasized the superior predictive power of Support Vector Machines (SVM), particularly with a radial kernel, achieving accuracies ranging from 89.4% to 91.8%. However, the SVM model experienced increased errors during the COVID-19 period, reflecting higher volatility and uncertainty in cryptocurrency returns influenced by the pandemic.

Similarly, Basher and Sadorsky (2022) conducted a study investigating the impact of macroeconomic news on the volatility of cryptocurrency returns, specifically focusing on Bitcoin and gold. The research utilized random forest models to predict the direction of Bitcoin and gold prices. Results indicated that random forests, tuned random forests, and tree bagging models outperformed logistic regression and boosted logit models, achieving an accuracy range of 75% to 80% for a 5-day forecast horizon for Bitcoin, with similar accuracies observed for gold. Furthermore, for a 15-day forecast horizon, the accuracy remained high, surpassing 90% for Bitcoin.The random forests models also demonstrated impressive accuracy exceeding 86% when considering a 20-day forecast horizon for both Bitcoin and gold. Time-series cross-validation further validated these high accuracies, with values ranging from 71% to 87% for Bitcoin and 63% to 88% for gold across various forecast horizons. Todorovska et al. (2023) conducted a study focusing on understanding the interconnectedness between classical economic indicators and cryptocurrencies through the application of machine learning (ML) and Explainable AI. The research identified Ethereum, Litecoin, Cardano, and Bitcoin as significant players in the cryptocurrency domain, while Dow Jones, S&P500, and FTSE were highlighted as crucial in traditional markets. Notably, in networks established using daily Twitter sentiment correlations, Chainlink and FTSE exhibited notable node degrees, with Chainlink leading in eigenvector centrality (0.47) and FTSE in node strength (0.31). Furthermore, in networks incorporating Explainable ML values, Chainlink, Stellar, Oil, and Price indicators demonstrated substantial influence, with Chainlink having the highest out-degree (5) and closeness centrality (327.13). These findings offer valuable insights into the intricate relationships between cryptocurrency and conventional financial markets, shedding light on market dynamics. Nguyen, Crane, and Bezbradica(2022) and Zahid, Iqbal, and Koutmos (2022) delved into the effectiveness of advanced machine learning techniques in predicting cryptocurrency volatility. Nguyen, Crane, and Bezbradica(2022) conducted a study on predicting the Cryptocurrency Volatility Index (CVI) using various models, with the AT-LSTM-MLP model emerging as the most effective. The results showed that the AT-LSTM-MLP model outperformed LSTM, SVR, RF, and TCN models, with significantly lower RMSE, MAE, and SMAPE values of 2.24 ± 0.17, 1.62 ± 0.09, and 1.76 ± 0.10, respectively. LSTM, the closest competitor, had errors more than double those of AT-LSTM-MLP.On the other hand, SVR, RF, and TCN models exhibited poorer results, showing significant deviations between predicted and real values. The study also evaluated Simple Moving Average (SMA) with different window sizes, revealing that while SMA performed well with smaller window sizes, its accuracy decreased as the window size increased. Specifically, SMA with a window size of 2 produced the best results, with MAE, RMSE, and SMAPE values of 2.02, 2.63, and 2.16, respectively. These findings highlight the effectiveness of the AT-LSTM-MLP model in accurately forecasting the CVI, providing valuable insights for cryptocurrency market analysis and risk management.

Similarly, Zahid, Iqbal, and Koutmos (2022) conducted a study on forecasting Bitcoin volatility using hybrid GARCH models combined with machine learning techniques. Analyzing Bitcoin price data from Jan 2015 to Mar 2021, the study observed significant deviations from normality and stationarity, prompting the need for advanced modeling approaches. GARCH-type models, including GARCH, GJR, and EGARCH, were applied, with EGARCH showing the lowest errors. Hybrid models, which integrated GARCH forecasts with deep learning (DL) algorithms such as LSTM, GRU, and BiLSTM, exhibited improved forecasting accuracy. The EGARCH-LSTM2 model emerged as the top performer, achieving relative improvements of up to 5.58% in HMAE and 6.06% in HMSE compared to single GARCH models. Combining double and triple GARCH-type models with DL models further enhanced accuracy, with the GARCH-GJR-EGARCH-LSTM2 model demonstrating significant improvements of up to 22.86% in HMAE and 33.92% in HMSE over single GARCH models. The rolling window approach also contributed to improved performance by reducing errors and enhancing forecasting accuracy. Through the MCS procedure, statistically significant models were identified, with the GARCH-GJR-EGARCH-LSTM2 model standing out as the best performer for Bitcoin volatility forecasting. These findings offer valuable insights for risk management and investment strategies in cryptocurrency markets.

Contrastingly, Ni, Härdle, and Xie (2020) conducted a study on the effectiveness of machine learning techniques in analyzing cryptocurrency market data and constructing a Regulatory Risk Index. The empirical results revealed that preprocessing steps, such as stopword elimination and lowercase conversion, improved the quality of the data for analysis. Using Latent Dirichlet Allocation (LDA) for topic modeling, the coherence value peaked at K = 14 topics, indicating optimal performance. The LDA model also exhibited topic diversification, with a low correlation between topics, as demonstrated by Jaccard and Hellinger distance matrices. In classification tasks, LDA outperformed Naive

Bayes and Support Vector Machine (SVM) methods, achieving an accuracy of 0.91 compared to 0.87 for the other two methods. Despite a higher Type I error rate, LDA excelled in identifying policy-related news, contributing effectively to index construction. The regulatory risk index derived from LDA-based classification is closely correlated with VCRIX, a volatility index for cryptocurrency markets, indicating its potential for forecasting market movements. Granger causality tests further confirmed the influence of regulatory risk (CRRIX) on market volatility (VCRIX), validating the predictive power of the regulatory risk index. The study highlights the efficacy of machine learning techniques, particularly LDA, in analyzing cryptocurrency market dynamics and informing risk management strategies.

# 6 Limitations

In reviewing the literature pertaining to the prediction of cryptocurrency prices, several studies have identified notable limitations, which collectively underscore the challenges inherent in this domain.

Several studies provided insights into the factors driving Bitcoin values, particularly focusing on the influence of social media. The reliance on Twitter sentiment as a proxy for market sentiment may not fully capture investor emotions (Georgoula et al. 2015) as it often reflects positive market patterns and persistent positive sentiment(Abraham et al. 2018). Moreover, the models sometimes struggle to accurately capture sentiment scores due to differences between real and synthetic tweets(Wong 2021). Studies employing sentiment analysis to model Bitcoin prices acknowledged limitations in their models' effectiveness with historical data (Gurrib and Kamalov 2022; Raju and Tarif 2020) and noted potential biases(Raju and Tarif 2020) or inconsistent improvements when using social media sentiment (Gurrib and Kamalov 2022). Additionally, the hybrid approach of sentiment analysis and machine learning models for predicting Bitcoin price fluctuations noted limitations in the quality of social signals from Twitter and the potential overlook of key information from other social networks(Valencia, Gómez-Espinosa, and Valdés-Aguirre2019).

Additionally, the use of OLS regressions might obscure nonlinear interactions between variables, and the small dataset employed may not capture the complex dynamics of Bitcoin markets adequately. Blockchain data alone is insufficient for predicting Bitcoin price changes, highlighting the need for external factors to create effective prediction models (Greaves and Au 2015). Wolk (2020) emphasized the impact of social media sentiment and web analytics on cryptocurrency prices but underscored the challenge of predicting prices due to market unpredictability. Singh and Bhat (2024) observed strong correlations among price, volume, and sentiments in cryptocurrencies but acknowledges the limited predictive capability of sentiment data due to insufficient large datasets. Moudhich and Fennan (2024) discussed the limitations of graph embedding techniques, particularly in scalability and handling out-of-vocabulary words, while Jung et al. (2023) highlighted the challenges of lexicon-based unsupervised learning for trend prediction, such as lower accuracy and inability to forecast magnitude of fluctuations.Şaşmaz and Tek (2021) also examined sentiment analysis but found varied accuracy between classifiers and restricted their focus to NEO cryptocurrency, limiting generalizability. Bao (2022) raised uncertainties regarding the real-world application of NLP and tspDB beyond predicting Bitcoin prices, necessitating further investigation. Uras and Ortu (2021) discussed the broader challenges in researching the cryptocurrency market, including its youthfulness, volatility, and technological complexities. El Abaji and A Haraty (2024) revealed that LSTM models, even with sentiment analysis, exhibit a conservative bias in long-term forecasts, consistent with some prior research but conflicting with others. This highlights the intricate relationship between historical price, sentiment analysis, and model architectures, offering insights for refining methodologies in financial forecasting, particularly in cryptocurrencies.

Finally, Ghosh, Jana, and D. K. Sharma (2024) focused solely on modeling the daily closing prices of five specific cryptocurrencies, potentially limiting the applicability of the findings to other cryptocurrencies or different periods and overlooking important variations in price behavior.

# 7 Future Recommendations

The burgeoning interest in cryptocurrencies has sparked a growing body of research aimed at understanding and predicting their price dynamics. Cryptocurrency price prediction holds significant implications for investors, traders, and policymakers alike, necessitating the development of robust forecasting models. Recent literature has highlighted various avenues for enhancing the accuracy and applicability of these models, ranging from advanced modeling techniques to sentiment analysis and behavioral analysis. In this chapter, we synthesized recent research findings and categorize future recommendations into distinct subtopics. We explore the potential directions for future research in cryptocurrency price prediction, emphasizing the importance of advanced modeling techniques, sentiment analysis, model performance enhancement, and market dynamics analysis. By delineating these future directions, we aim to provide a comprehensive roadmap for researchers and practitioners in cryptocurrency price prediction.

## 7.1 Advanced Modeling Techniques and Performance Enhancement

Recent studies have underscored the significance of advanced modeling techniques and comprehensive data integration in enhancing the accuracy of cryptocurrency price prediction models. Georgoula et al. (2015) advocated for the adoption of advanced econometric tools like vector autoregressive models to gain a nuanced understanding of short-term dynamics. Additionally, Abraham et al. (2018) suggested exploring machine learning algorithms beyond linear regression to unlock further gains in predictive accuracy. Valencia, Gómez-Espinosa, and Valdés-Aguirre(2019) emphasized the potential of specialized models incorporating attention mechanisms, such as long short-term memory networks (LSTM) and temporal multi-layer perceptrons (T-MLP) in capturing market sentiment more effectively. Jagannath et al. (2021) proposed improving deep learning models through the integration of supplementary self-adaptive techniques and correlating them with the behavior and price of crypto assets aims to boost prediction accuracy for Bitcoin and other cryptocurrencies. Gurrib and Kamalov (2022) proposed extending research to include other cryptocurrencies to examine model applicability across digital assets and also incorporating more diversified classifiers, like decision trees and long short-term memory networks. Onyekwere, Ogwueleka, and Irhebhude (2022)and Othman et al. (2020) suggested that future research should expand to compare several types of artificial neural network models for forecasting the long-term viability of blockchain technology. Parekh et al. (2022) highlighted the future direction of developing unified forecasting models with advanced architectures like transformers and federated learning models, capable of efficiently predicting prices for various cryptocurrencies while integrating additional factors such as sentiment analysis and blockchain characteristics.

Enhancing model performance through refined data preprocessing methods and alternative machine learning approaches remains a focal point for future research. Greaves and Au (2015) highlighted the importance of including Bitcoin exchange-specific aspects, leveraging second-by-second exchange data, analyzing user adoption and activity within exchanges, and exploring unique exchange characteristics to enhance Bitcoin price prediction model accuracy. McCoy and Rahimi (2020) proposed experimenting with alternative machine learning methods and refining training strategies to optimize model performance. Wong (2021) suggested further refining data preprocessing methods and mitigating human bias to enhance predictive accuracy. Moreover, future studies should prioritize improving model performance by incorporating exchange-specific features and refining data preprocessing techniques to produce more robust and accurate prediction models as stated by Cheng et al. (2024). Raju and Tarif (2020) highlighted enhancing model performance by incorporating real-time streaming data into predictive analysis and investigating the integration of additional social media platforms like Facebook and LinkedIn to gain a comprehensive understanding of public sentiment.

Moreover, merging current sentiments with LSTM projections could enable automated trading assistants to make informed decisions regarding Bitcoin transactions. Lamon, Nielsen, and Redondo (2017) suggested refinement through experimentation and updates, including training with a blend of news and Twitter data to enhance classification and prediction robustness across diverse trends. Additionally, exploration will extend to integrating supplementary media sources and larger datasets, experimenting with varied strategies for labeling training data, and improving text pre-processing and feature consideration in the baseline model. Moudhich and Fennan (2024) advocated integrating graph embeddings with attention mechanisms and transformers while also exploring novel graph construction methods like knowledge graph integration to enhance model effectiveness and sentiment analysis accuracy in the future. Ghosh, Jana, and D. K. Sharma (2024) recommended exploring the predictability of cryptocurrency prices across different time intervals, including hourly, 5-minute, and 1-minute intervals, to support trading strategies at varied time horizons, with the potential for assessing cryptocurrency index predictability. Bâra and Oprea (2024) advised including comprehensive datasets comprising 1-minute numerical records and textual data from social media platforms and news outlets, suggesting the integration of transfer learning with quantum computing for accelerated processing, alongside incorporating blockchain analytics and on-chain metrics to gain deeper insights into market dynamics.

## 7.2 Sentiment Analysis and Social Media Integration

The integration of sentiment analysis from diverse sources such as social media platforms and news outlets emerges as a pivotal area for future research. Wu (2023) suggested exploring different sentiment analysis methods like FinBERT or Naive Bayes, alongside ensemble learning boosting methods such as XGBoost or LightGBM, while also expanding the dataset could provide a more accurate representation of sentiment.Singh and Bhat (2024) advocated for exploring sentiment data from diverse sources like news articles and leveraging Reddit posts and comments comprehensively, while also exploring existing models like TST, Autoformer, and TimeGPT-1, alongside developing new time series transformers to capture data interactions more effectively. Similarly,Mirtaheri et al. (2021) proposed enhancing predictive capabilities by expanding datasets to include Reddit posts, developing a specialized bot detection system for cryptocurrencies, and implementing a real-time monitoring system to detect pump attacks and alert vulnerable users. Lastly, Poongodi et al. (2021) suggested including Ethereum and Litecoin, expanding datasets with more tweets and Reddit posts, and testing models on extensive data from diverse forums and social platforms to validate social media trends as robust cryptocurrency indicators. According to Şaşmaz and Tek (2021), future studies may enhance sentiment analysis by augmenting Random Forest model training data, refining BERT-based models with cryptocurrency-specific data, and broadening research scope to encompass sentiments across multiple cryptocurrencies for a comprehensive understanding of sentiment-price relationships in the cryptocurrency market. Raheman et al. (2022) advised to automate the utilization of price movements as implicit tags for sentiment-rich text data, leveraging indicative n-grams learned from temporally aligned market and news media data, with an emphasis on enhancing price prediction accuracy, particularly in decentralized finance applications. Haritha and Sahana (2023) suggested integrating sentiments from diverse sources like mainstream news, online articles, Reddit, and Twitter to forecast crypto prices. El Abaji and A Haraty (2024) recommended investigating the incorporation of predictive models and sentiment analysis into portfolio optimization frameworks, especially within cryptocurrency markets. This integration offers the opportunity to augment conventional portfolio strategies by incorporating market sentiments alongside historical data, thereby potentially enhancing risk management and investment decision processes. However, challenges associated with handling volatility in sentiment data and navigating the speculative characteristics inherent in cryptocurrency markets necessitate thorough consideration and the implementation of robust risk management measures.

### 7.3 Market Dynamics and Behavioral Analysis

Understanding market dynamics and behavioral factors influencing cryptocurrency prices remains a crucial area for future research. Inamdar et al. (2019) emphasized the importance of incorporating supplementary factors such as news sources and blockchain characteristics to improve price predictions for extended periods. El Haddaoui et al. (2023) proposed enriching classified data with sources like Reddit and news outlets to provide a more accurate portrayal of public sentiment towards cryptocurrencies. Carbó and Gorjón(2024) suggested examining factors affecting Bitcoin's price, including the influence of trading platforms and inter-exchange price interactions, alongside socio-technological aspects like impactful tweets and the impact of bubbles and herd behavior, could provide valuable insights into the driving forces behind Bitcoin's price fluctuations. Kim, Trimborn, and Härdle (2021) mentioned to conduct further research to comprehend the substantial volatility observed in derivative-based indices like VIX, influenced by behavioral aspects of option pricing.Jain et al. (2018) suggested social criteria such as user credibility, popularity, and network can be used to accurately anticipate bitcoin prices. Lamothe-Fern´andez et al. (2020) advocated the development of a predictive model considering volatility correlations among emerging alternative assets and safe-haven assets such as gold or stable currencies, evaluating diverse portfolio options and optimization scenarios.

## 8 Conclusion

Cryptocurrency price prediction stands at the intersection of finance, data science, behavioral economics, and industrial engineering, captivating the attention of researchers, investors, policymakers, and engineers. This chapter has navigated through a landscape marked by both remarkable advancements and persistent challenges, offering valuable insights into the current state of research in this dynamic field.

Our evaluation of various modeling approaches in cryptocurrency price prediction reveals the evolution and complexity of the techniques employed. While linear regression models provide a foundational framework, they often falter in capturing the intricate dynamics inherent in cryptocurrency markets. Advanced techniques such as decision trees and support vector machines offer enhanced flexibility, yet struggle to adapt to the high dimensionality and volatility of the data. On the other hand, deep learning architectures, notably convolutional neural networks (CNNs) and long short-term memory networks (LSTMs), demonstrate superior performance in feature extraction and sentiment capture, albeit with computational complexities and interpretability challenges. Industrial engineers can play a critical role in optimizing these models, particularly by applying operations research techniques to enhance computational efficiency and data management strategies.

Despite these advancements, several limitations persist. Reliance on historical data, small dataset sizes, and the failure to adequately adapt to evolving market dynamics remain prominent challenges. Additionally, issues related to interpretability and bias from sentiment analysis on social media sources highlight the need for caution in model interpretation and decision-making. Industrial engineers, with their expertise in systems optimization, can contribute significantly to overcoming these challenges by designing more robust data pipelines and leveraging techniques such as multi-task learning to handle diverse data types effectively.

Future research directions aim to address these limitations and enhance the accuracy and robustness of cryptocurrency price prediction models. The integration of advanced modeling techniques such as transformers and federated learning models holds promise for efficiently predicting prices across various cryptocurrencies while incorporating additional factors like sentiment analysis and blockchain characteristics. With their experience in system design and process optimization, industrial engineers are well-positioned to contribute to developing these advanced models, ensuring that scalability and efficiency are maintained even in the face of large, complex datasets. Moreover, the augmentation of sentiment analysis with diverse data sources and the exploration of alternative machine learning approaches aim to refine model performance and improve predictive capabilities.

In conclusion, while predicting cryptocurrency prices is fraught with challenges, the insights gleaned from this chapter serve as signposts guiding future research endeavors. By embracing the complexities of the cryptocurrency market and leveraging emerging technologies, including the contributions of industrial engineers, researchers can pave the way for more informed investment decisions, robust regulatory policies, and a deeper understanding of this transformative asset class.